\documentclass[runningheads]{llncs}

\usepackage[mobile]{eccv}

\usepackage{eccvabbrv}

\usepackage{graphicx}
\usepackage{booktabs}
\usepackage{multirow}
\usepackage{threeparttable}
\usepackage{tabularx}
\usepackage{ragged2e}
\usepackage{microtype}
\usepackage[ruled, linesnumbered, boxed]{algorithm2e} 
\usepackage{pifont}
\usepackage{amsmath}

\usepackage[accsupp]{axessibility}  

\usepackage[
    colorlinks=true,
    linkcolor=magenta,
    urlcolor=magenta,
    citecolor=eccvblue
]{hyperref}

\usepackage{orcidlink}

\begin{document}

\title{AutoWeather4D: Autonomous Driving Video Weather Conversion via G-Buffer Dual-Pass Editing} 

\titlerunning{AutoWeather4D}

\author{
\centering
Tianyu Liu\inst{1*} \quad
Weitao Xiong\inst{1,2*} \quad
Kunming Luo\inst{1} \quad \\[0.25em]
Manyuan Zhang\inst{3} \quad 
Peng Li\inst{1} \quad
Yuan Liu\inst{1\dagger} \quad
Ping Tan\inst{1\dagger}\\[0.5em]
\small \url{https://lty2226262.github.io/autoweather4d}\vspace{-0.5em}
}

\authorrunning{T. Liu, W. Xiong et al.}

\institute{
\centering
$^{1}$The Hong Kong University of Science and Technology \quad \\[0.25em]
$^{2}$Xiamen University\quad $^{3}$Meituan-M17, Hong Kong\\[0.5em]
$^{*}$Equal Contribution\quad
$^{\dagger}$Equal Corresponding Author
}

\maketitle

\definecolor{myred}{RGB}{180,23,1}
\definecolor{myblue}{RGB}{0,118,186}

\begin{figure}[t]
  \centering
    \includegraphics[width=0.95\textwidth]{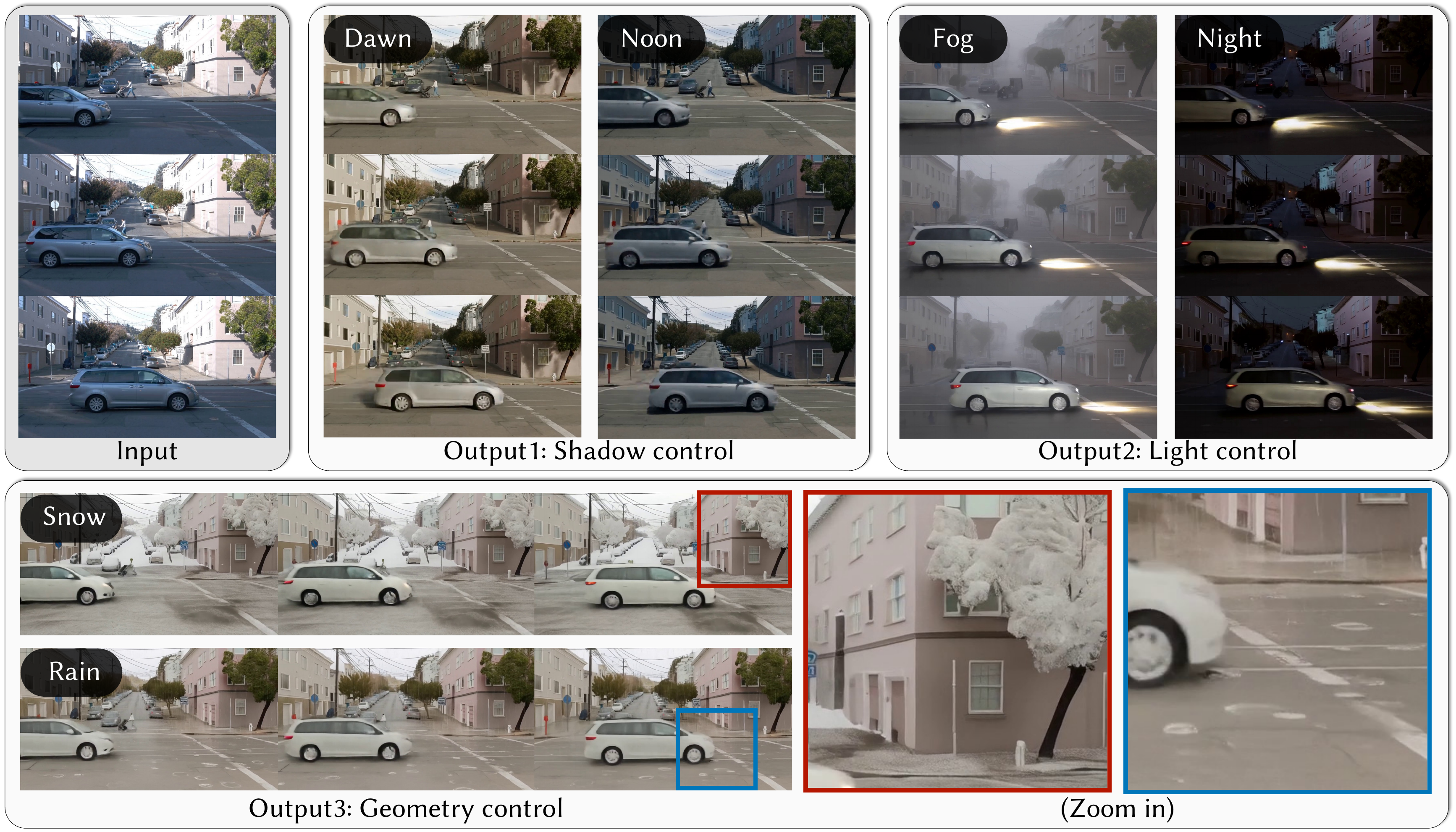}
    \caption{\textbf{AutoWeather4D: Weather \& Time-of-Day Control for Driving Videos}. AutoWeather4D enables fine-grained control over weather (rain, snow, fog) and time-of-day (dawn, noon, night) in driving videos. The \textcolor{myred}{red zoom-in box} showcases realistic snow accumulation, while the \textcolor{myblue}{blue zoom-in box} highlights rain-induced road wetness and ripples. See supplementary videos for dynamic visualizations.}
    \label{fig:teaser_results}
    \vspace{-2em}
\end{figure}

\begin{abstract}

Generative video models have significantly advanced the photorealistic synthesis of adverse weather for autonomous driving; however, they consistently demand massive datasets to learn rare weather scenarios. While 3D-aware editing methods alleviate these data constraints by augmenting existing video footage, they are fundamentally bottlenecked by costly per-scene optimization and suffer from inherent geometric and illumination entanglement. In this work, we introduce AutoWeather4D, a feed-forward 3D-aware weather editing framework designed to explicitly decouple geometry and illumination. At the core of our approach is a G-buffer Dual-pass Editing mechanism. The Geometry Pass leverages explicit structural foundations to enable surface-anchored physical interactions, while the Light Pass analytically resolves light transport, accumulating the contributions of local illuminants into the global illumination to enable dynamic 3D local relighting. Extensive experiments demonstrate that AutoWeather4D achieves comparable photorealism and structural consistency to generative baselines while enabling fine-grained parametric physical control, serving as a practical data engine for autonomous driving.

\end{abstract}
\section{Introduction}
\label{sec:intro}

Recent advances in generative video models~\cite{bai2025ditto, lin2025controllableweathersynthesisremoval, zhu2025scenecrafter, zhu2025weatherdiffusionweatherguideddiffusionmodel, wan2025, nvidia2025worldsimulationvideofoundation} represent an important step towards the photorealistic synthesis of adverse weather conditions for autonomous driving. However, despite their impressive visual fidelity, these data-driven approaches consistently demand massive datasets to learn rare adverse weather patterns. Capturing such long-tail environmental data in the real world remains prohibitively expensive and logistically constrained.

To circumvent these data constraints, 3D-aware editing methods~\cite{Li2023ClimateNeRF, dai2025rainygs, weatheredit, weathermagician} offer a compelling alternative by augmenting existing video footage. By explicitly grounding the synthesis process in 3D space, these approaches achieve high-fidelity and highly controllable weather effects without relying on massive, long-tail training datasets. They typically operate through a straightforward two-stage pipeline: first reconstructing a 3D representation of the captured scene, and subsequently applying weather-specific modifications to the underlying geometry and appearance. However, these methods are fundamentally bottlenecked by their reliance on painstakingly slow per-scene optimization. Requiring up to an hour of computation per video clip, this optimization paradigm is computationally prohibitive for large-scale data generation. 

In this paper, we propose a 3D-aware editing method called AutoWeather4D, which brings the controllability and visual quality of 3D-aware editing to dynamic autonomous driving scenarios. By replacing the sluggish per-scene optimization with a novel feed-forward editing pipeline that explicitly decouples geometry and illumination, AutoWeather4D achieves rapid, high-quality, and physically plausible weather editing and lighting editing.

Designing such a framework is non-trivial, which first requires defining an editable and flexible 3D scene representation for the dynamic autonomous driving scenes. Existing 3D-aware editing pipelines heavily rely on scene representations like NeRF~\cite{nerf} or 3DGS~\cite{kerbl3Dgaussians}. However, a fundamental limitation of these frameworks is their inherent reliance on static scene assumptions for high-quality reconstruction. When confronted with the complex, highly dynamic environments typical of autonomous driving with moving vehicles and pedestrians, these optimized fields frequently fail to capture accurate underlying 3D geometry. Consequently, spatially anchoring and consistently applying weather effects across dynamic elements becomes exceptionally difficult. 

To address this, our method represents the dynamic scene by the extracted G-buffers of the videos with a feed-forward neural network~\cite{DiffusionRenderer}. By directly predicting dense, frame-wise geometric features (such as depth and normals) from the video stream, we entirely bypass the static-scene bottleneck of per-scene optimization. This explicit G-buffer formulation natively accommodates dynamic objects and provides a highly controllable, reliable structural foundation, making downstream weather editing both intuitive and geometrically precise.

Second, building upon our explicit geometric foundation, we address the severe illumination entanglement that plagues current weather editing paradigms. Existing 3D-aware methods typically assume a static, single global illumination setup, fundamentally baking the original scene's appearance and lighting directly into the optimized 3D representation. While this may suffice for static landscapes, it completely breaks down in dynamic autonomous driving environments. In these complex scenarios, realistic weather synthesis inherently requires modeling dynamic local lighting, such as moving vehicle headlights sweeping across wet surfaces or streetlights creating volumetric halos in the fog. To break this architectural barrier, AutoWeather4D introduces a fully decoupled Light Pass. By integrating physics-based lighting priors with our G-buffer-driven neural rendering, our framework thoroughly separates the global atmospheric conditions from localized, dynamic illuminants. This explicit decoupling unlocks the unprecedented ability to seamlessly insert, toggle, and physically relight 3D local sources under adverse weather conditions, ensuring that both static and dynamic elements react accurately to environmental changes (See Fig.~\ref{fig:teaser_results}).

\begin{table*}[t]
    \centering
    \resizebox{0.98\textwidth}{!}{ 
        \begin{tabular}{lccccccc}
            \toprule  
            Method & 
            Env light/shadow control & 
            Extra light source & 
            Weather change & 
            Feed-forward & 
            4D dynamic scene & 
            Tuning-free & 
            Open-source \\
            \midrule  
            Cosmos-Transfer2.5~\cite{nvidia2025worldsimulationvideofoundation} & {\color{red}$\times$} & {\color{red}$\times$} & {\color{ForestGreen}\checkmark} & {\color{ForestGreen}\checkmark} & {\color{ForestGreen}\checkmark} & {\color{ForestGreen}\checkmark} & {\color{ForestGreen}\checkmark} \\
            WAN-FUN 2.2~\cite{wan2025} & {\color{red}$\times$} & {\color{red}$\times$} & {\color{ForestGreen}\checkmark} & {\color{ForestGreen}\checkmark} & {\color{ForestGreen}\checkmark} & {\color{ForestGreen}\checkmark} & {\color{ForestGreen}\checkmark} \\
            Ditto~\cite{bai2025ditto} & {\color{red}$\times$} & {\color{red}$\times$} & {\color{ForestGreen}\checkmark} & {\color{ForestGreen}\checkmark} & {\color{ForestGreen}\checkmark} & {\color{ForestGreen}\checkmark} & {\color{ForestGreen}\checkmark} \\
            WeatherWeaver~\cite{lin2025controllableweathersynthesisremoval} & {\color{red}$\times$} & {\color{red}$\times$} & {\color{ForestGreen}\checkmark} & {\color{ForestGreen}\checkmark} & {\color{ForestGreen}\checkmark} & {\color{red}$\times$} & {\color{red}$\times$} \\
            WeatherDiffusion~\cite{zhu2025weatherdiffusionweatherguideddiffusionmodel} & {\color{red}$\times$} & {\color{red}$\times$} & {\color{ForestGreen}\checkmark} & {\color{ForestGreen}\checkmark} & {\color{red}$\times$} & {\color{red}$\times$} & {\color{red}$\times$} \\
            SceneCrafter~\cite{zhu2025scenecrafter} & {\color{red}$\times$} & {\color{red}$\times$} & {\color{ForestGreen}\checkmark} & {\color{ForestGreen}\checkmark} & {\color{ForestGreen}\checkmark} & {\color{red}$\times$} & {\color{red}$\times$} \\
            RainyGS~\cite{dai2025rainygs} & {\color{red}$\times$} & {\color{red}$\times$} & {\color{ForestGreen}\checkmark} & {\color{red}$\times$} & {\color{ForestGreen}\checkmark} & {\color{ForestGreen}\checkmark} & {\color{red}$\times$} \\
            WeatherEdit~\cite{weatheredit} & {\color{red}$\times$} & {\color{red}$\times$} & {\color{ForestGreen}\checkmark} & {\color{red}$\times$} & {\color{ForestGreen}\checkmark} & {\color{ForestGreen}\checkmark} & {\color{ForestGreen}\checkmark} \\
            ClimateNeRF~\cite{Li2023ClimateNeRF} & {\color{red}$\times$} & {\color{red}$\times$} & {\color{ForestGreen}\checkmark} & {\color{red}$\times$} & {\color{red}$\times$} & {\color{ForestGreen}\checkmark} & {\color{ForestGreen}\checkmark} \\
            DiffusionRenderer~\cite{DiffusionRenderer} & {\color{ForestGreen}\checkmark} & {\color{red}$\times$} & {\color{red}$\times$} & {\color{ForestGreen}\checkmark} & {\color{ForestGreen}\checkmark} & {\color{ForestGreen}\checkmark} & {\color{ForestGreen}\checkmark} \\
            \midrule
            AutoWeather4D(Ours) & {\color{ForestGreen}\checkmark} & {\color{ForestGreen}\checkmark} & {\color{ForestGreen}\checkmark} & {\color{ForestGreen}\checkmark} & {\color{ForestGreen}\checkmark} & {\color{ForestGreen}\checkmark} & {\color{ForestGreen}\checkmark} \\
            \bottomrule
        \end{tabular}
    }
    \caption{Comparison of state-of-the-art weather and time-of-day synthesis models for autonomous driving. We evaluate existing paradigms across key capabilities: Env light/shadow control (arbitrarily adjusting environment light directions and correcting shadows), Extra light source (precisely adding and controlling local lights within driving scenes), and Weather change (synthesizing appearances under diverse weather conditions). Additionally, we compare their architectural and practical properties, including whether they are Feed-forward (requiring no per-scene optimization), applicable to 4D dynamic scenes, Tuning-free (requiring no extra datasets for weight fine-tuning). }
    \label{tab:comparison} 
    \vspace{-2em}

\end{table*}

Extensive experiments on standard autonomous driving datasets demonstrate that AutoWeather4D synthesizes adverse weather and illumination conditions from existing footage, without requiring any auxiliary data. As summarized in Tab.~\ref{tab:comparison}, compared to existing paradigms, our framework uniquely achieves decoupled control over geometric weather elements and global/local light transport without the need for per-scene optimization.

In summary, our main contributions are:

\begin{itemize}
\item We introduce AutoWeather4D, a feed-forward 3D-aware weather editing framework. It synthesizes adverse weather conditions from real-world driving videos while eliminating the need for per-scene optimization.

\item We propose a G-buffer Dual-pass Editing mechanism. The Geometry Pass enables surface-anchored interactions (e.g., snow accumulation), while the Light Pass analytically accumulates local illuminants for 3D relighting.

\item Experiments demonstrate that AutoWeather4D achieves comparable photorealism and structural consistency to generative baselines while enabling parametric physical control, serving as a practical data engine for autonomous driving.
\end{itemize}

\vspace{-1em}
\section{Related Works}
\label{sec:related_work}
\vspace{-0.5em}
In this section, we review two related topics. First, we
discuss advances in Climate simulation. Next, we
discuss video simulator for autonomous driving.
\vspace{-1em}
\subsection{Climate simulation}
\vspace{-0.5em}


\textbf{Physical based simulator:} Classical computer graphics have long established the physical foundations for rendering weather effects like rain~\cite{particle_system,rain_texture,realtime_rain}, snow~\cite{metaball,realtime_snow,snow_opengl}, and volumetric fog~\cite{fog_scatter,metaball_fog,realtime_fog} using particle systems and scattering equations. However, these classical methods fundamentally rely on explicit 3D meshes or voxel grids and cannot be directly applied to monocular videos. Our method seamlessly integrates these classical physical priors into real-world video footage, preserving physical guarantees while enabling flexible video-based editing.

\textbf{Network-based simulator:} The advent of deep learning has enabled data-driven approaches to climate and weather simulation. In the image domain, early work applied CycleGAN~\cite{cyclegan} for weather transfer~\cite{climate_cyclegan}, while diffusion models such as Prompt-to-Prompt~\cite{hertz2023prompt2prompt} and SDEdit~\cite{meng2022sdedit} enabled weather modification via text prompts or sketch-based guidance. Recent methods target illumination control specifically: LightIt~\cite{lightit} conditions generation on one-bounce shadow maps; Retinex-Diffusion~\cite{retinex-diffusion} reformulates the energy function of diffusion models to fulfill illumination alteration; DiLightNet~\cite{dilightnet} and IntrinsicAnything~\cite{intrinsicanything} decompose images into BRDF components for relighting; and IC-Light~\cite{iclight} adjusts illumination based on reference backgrounds. Video extensions include fine-tuning-based editors (WeatherWeaver~\cite{lin2025controllableweathersynthesisremoval}, WeatherDiffusion~\cite{zhu2025weatherdiffusionweatherguideddiffusionmodel}, SceneCrafter~\cite{zhu2025scenecrafter}, Ditto~\cite{bai2025ditto}), ControlNet-style conditioning methods (WAN-FUN 2.2~\cite{wan2025}, Cosmos-Transfer2.5~\cite{nvidia2025worldsimulationvideofoundation}), and G-buffer decomposition approaches (DiffusionRender~\cite{DiffusionRenderer}). However, existing methods address either weather conversion or physically-based illumination control, but not both simultaneously—a critical limitation that our work resolves through unified physical rendering and diffusion-based synthesis.

\textbf{Hybrid physics-and-learning simulators.} Recent methods integrate classical graphics with deep learning across diverse 3D representations. NeRF-based approaches \cite{Li2023ClimateNeRF} embed physical weather models or text-guided editing into neural radiance fields for high-fidelity rendering of atmospheric effects (fog, snow, flooding), though limited to static scenes. Mesh-based techniques~\cite{dreameditor, video2game} convert NeRF~\cite{nerf} reconstructions into interactive meshes with rigid-body physics for real-time interaction. Gaussian Splatting methods leverage 3DGS~\cite{kerbl3Dgaussians} for efficient rendering: GaussianEditor~\cite{gaussianeditor}) enable cross-view 2D-to-3DGS manipulation, RainyGS~\cite{dai2025rainygs} models physical raindrops, Weather-Magician~\cite{weathermagician} incorporates depth/normal supervision for multi-weather synthesis, DRAWER~\cite{drawer} combine 3DGS with mesh for articulated things, and WeatherEdit~\cite{weatheredit} extends to 4DGS for temporal control. Unlike these per-scene optimization reconstruction approaches, our method employs feed-forward 4D reconstruction~\cite{dust3r,vggt,pi3}, which—despite producing sparser outputs that pose additional challenges—eliminates scene-specific tuning and drastically reduces deployment time.

\vspace{-1em}
\subsection{Autonomous driving video simulator}
\vspace{-0.5em}

Autonomous driving world model simulators~\cite{magicdrive, vista, gaia2, drivingdiffusion, longvideogeneration, drivedreamer4d, wei2024editable, unisim, panacea, zhu2025scenecrafter, causnvs, occsora, r3d2, adriveri, lightsim, recondreamer} play a crucial role in generating complex traffic scenarios that are challenging to capture in real-world conditions, substantially reducing data collection costs for training self-driving systems—particularly benefiting end-to-end autonomous driving. Unlike prior approaches that rely on iterative neural optimization or latent-space manipulation, our method leverages classical graphics techniques by directly operating on the G-buffer, enabling explicit geometric and illumination control for efficient scene modification, which is absent in existing simulators.

\vspace{-1em}
\section{Method}
\vspace{-0.5em}
\label{sec:method}
\begin{figure*}[t]
  \centering
    \includegraphics[width=1.0\linewidth]{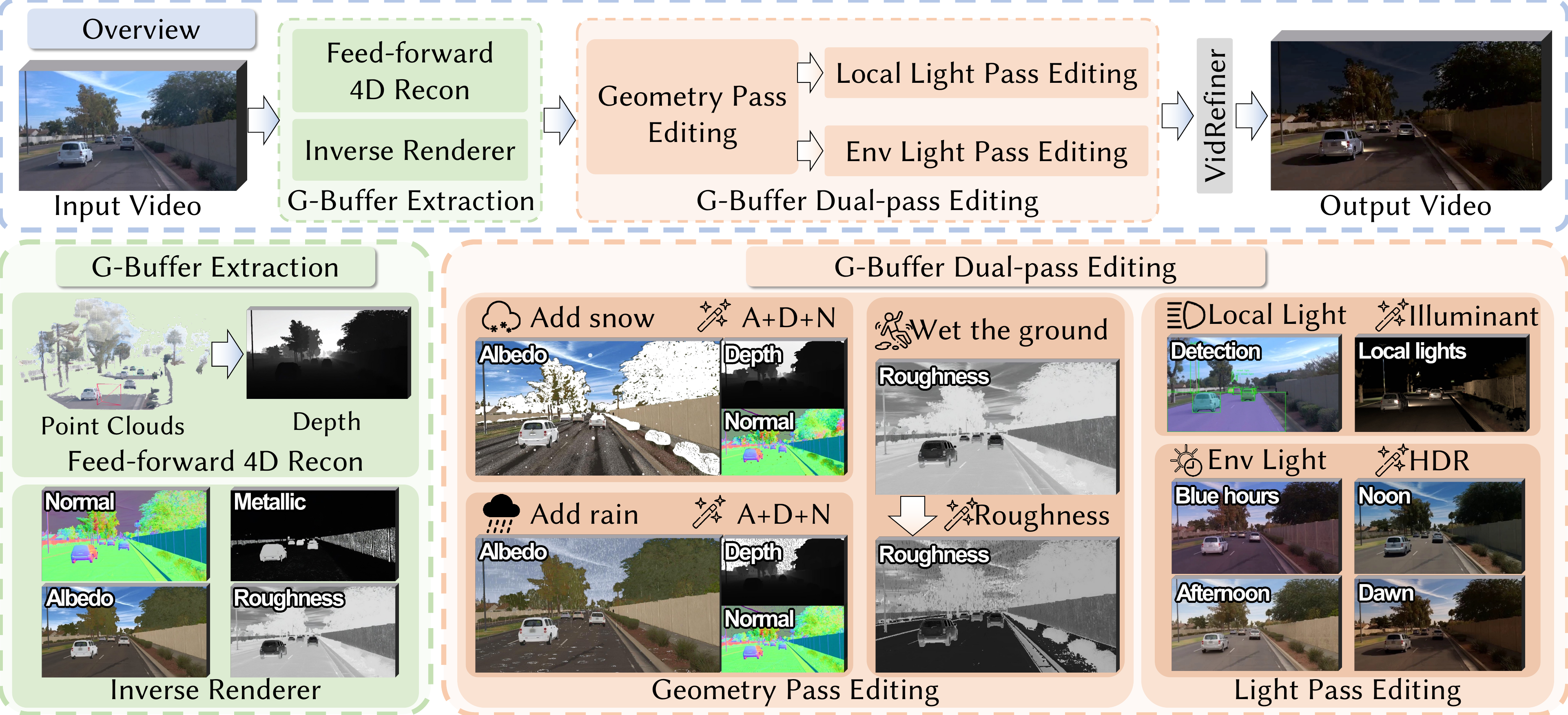}
    \caption{\textbf{Overview of our framework.} The pipeline formulates physically-grounded video editing for multi-weather and time-of-day synthesis. We first extract explicit G-buffers from the input video: metric depth $\mathbf{D}$ via feed-forward 4D reconstruction, alongside intrinsic material properties (normal $\mathbf{N}$, metallic $\mathbf{M}$, albedo $\mathbf{A}$, roughness $\mathbf{R}$) via an inverse renderer. The scene modifications are analytically resolved through the \textbf{G-Buffer Dual-Pass Editing}: (1) The \textit{Geometry Pass} physically modulates $\mathbf{A, N, R}$ to instantiate explicit weather mechanics (e.g., snow, rain, ground wetness); (2) The \textit{Light Pass} executes parametric illumination control, independently synthesizing detected local light sources and global environmental lighting (e.g., dawn, noon, blue hours) to reflect atmospheric and temporal shifts. Finally, the deterministic rendered sequence is processed by the \textbf{VidRefiner}. This terminal refiner synthesizes real-world sensor nuances while preserving the classical shading cues and explicit scene dynamics resolved in the dual-pass stages.
}
\vspace{-2em}
    \label{fig:main}
\end{figure*}

\textbf{Overview.} As shown in Fig.~\ref{fig:main}, we formulate weather editing as an efficient \textit{analysis-and-synthesis} pipeline. The \textit{analysis} stage decomposes the input video into explicit intrinsic G-buffers (Sec.~\ref{sec:4d recon}) in a feed-forward manner, bypassing the prohibitive cost of per-scene optimization. The subsequent \textit{synthesis} stage manipulates the decoupled scene geometry and illumination via a Dual-pass Editing mechanism (Sec.~\ref{sec: g_buffer_dual_pass_editing}). Finally, the VidRefiner performs terminal refinement on the rendered sequence, incorporating sensor nuances while conditioning the generative process on the resolved physical dynamics (Sec.~\ref{sec:videorefiner}).

\vspace{-1em}
\subsection{Feed-Forward G-Buffer Extraction}
\label{sec:4d recon}
\vspace{-0.5em}

\textbf{Feed-forward Intrinsic Parsing.} To bypass the static-scene assumptions of implicit optimization and ensure accurate geometric anchoring in dynamic environments, the monocular sequence is parsed into a unified G-buffer through a multi-source feed-forward extraction scheme. Initially, spatiotemporally coherent relative depth is instantiated by deploying Pi3~\cite{pi3}, a feed-forward 4D reconstruction backbone. Alongside this geometric extraction, intrinsic material properties (albedo, normal, metallic, roughness) are decoupled via a zero-shot diffusion-based inverse renderer~\cite{DiffusionRenderer}. Consolidating these multi-source extractions yields a preliminary state for downstream editing processing, necessitating absolute metric scale depth and spatial bounding to ensure physical validity.

\textbf{Relative Depth Alignment.} The relative scale of the reconstructed geometry fundamentally conflicts with the absolute metric requirements of physical light transport. To establish an absolute physical scale, the global scalar multiplier is deterministically resolved by aligning the relative depth with sparse LiDAR point clouds. For strictly monocular configurations lacking LiDAR, this scaling factor is alternatively recovered via standard geometric priors, such as known camera height~\cite{cameraheight}. This calibration maintains framework adaptability while ensuring exact metric alignment for subsequent editing and relighting mechanics.

\textbf{Sky-Aware Material Extraction.} Furthermore, to prevent artifacts from infinite-depth regions during material estimation, we implement a dedicated sky-masking mechanism. This ensures that the diffusion-based material priors are strictly constrained to valid scene geometry, guaranteeing pixel-level correspondence and structural stability for downstream geometry and light manipulation.

The implementation details of alignment and sky-aware material extraction are provided in (Sec.~\ref{sec:4d_reconstruction_details} in supplementary materials). 

\vspace{-1em}
\subsection{G-Buffer Dual-pass Editing}
\label{sec: g_buffer_dual_pass_editing}

To extend high-fidelity 3D-aware editing to dynamic driving scenarios, we propose a Dual-Pass Editing mechanism. This pipeline systematically decouples structural scene modifications from illumination transport: the \textbf{Geometry Pass} first updates the intrinsic state of the scene, which then serves as the physical foundation for the \textbf{Light Pass} to analytically resolve radiance. By operating on explicit G-buffers, this mechanism ensures that all synthesized environmental changes remain anchored to the underlying 3D structure.

\subsubsection{Geometry Pass: Surface-Anchored Interaction}

The Geometry Pass transforms the intrinsic albedo, normal, and roughness to incorporate the physical presence of weather elements. These updated surface descriptors parameterize the subsequent Light Pass, ensuring all illumination transport is analytically resolved over the modified scene structure. Specifically, we instantiate these surface-anchored modifications through explicit physical models for two representative weather conditions:

\textbf{Multi-Representation Snow Synthesis.} To bridge the scale gap between individual snowflakes and terrain-scale coverage, we employ a hybrid simulation: (1) \textit{Metaball-based Surface Buildup} iteratively evaluates an SPH Poly6 kernel~\cite{10.5555/846276.846298} over the extracted normal maps, restricting accumulation to upward-facing structures to maintain geometric rational; (2) \textit{Grid-based Ground Modeling} utilizes procedural patterns for varied snow density alongside a physically-based wetness model that darkens albedo and reduces roughness to simulate thawing transitions; and (3) \textit{Kinematic Falling Particles} are rendered via temporally-persistent screen-space rasterization to ensure inter-frame kinematic continuity. Implementation details are provided in (Sec.~\ref{sec:snow_details} of the supplementary material).

\textbf{Physically-Grounded Rain Dynamics.} We decouple rain synthesis into kinematic streaks and standing water. Falling drops are modeled as kinematic particles governed by a vector summation of Gunn–Kinzer terminal velocities~\cite{1971JApMe..10..751W} (vertical gravity-drag equilibrium) and parametric wind fields (horizontal displacement). We parameterize these trajectories as volumetric Signed Distance Fields (SDFs), explicitly depth-testing against the extracted depth to enforce precise spatial occlusion. For ground interactions, puddle masks generated via Fractional Brownian Motion (FBM) physically modulate the local albedo and roughness. Concurrently, surface normals within these masked regions are perturbed using procedural ripple maps to approximate dynamic impact responses. Implementation details are provided in (Sec.~\ref{sec:rain_details} of the supplementary material).

\vspace{-1em}
\subsubsection{Light Pass: Decoupled Illumination Control}
Given the updated G-buffers from the Geometry Pass, the Light Pass computes the final scene illumination. By operating directly on these explicit material properties, we can independently synthesize local light sources and global atmospheric scattering, enabling direct parametric relighting. Specifically, we instantiate this parametric relighting through tailored physical models for three representative illumination scenarios:

\textbf{Nocturnal Local Relighting.} We explicitly model artificial sources (e.g., streetlights, headlights) as 3D spotlights, estimating their spatial positions via semantic masks and the metric depth. Surface radiance is then analytically evaluated using the Cook-Torrance BRDF~\cite{10.1145/357290.357293}, which is directly parameterized by the edited G-buffers to enforce physically consistent material responses. For non-illuminated regions, a parametric Look-Up Table (LUT) shifts ambient color temperatures toward warm nocturnal tones to maintain minimal visibility. Light sources estimation details are provided in (Sec.~\ref{sec: semantic annotations} of the supplementary materials), while the BRDF and LUT implementation details are provided in (Sec.~\ref{sec: night_details} of the supplementary materials).

\textbf{Volumetric Atmospheric Scattering.} We formulate foggy environments by analytically resolving volumetric scattering via a single-scattering Radiative Transfer Equation (RTE) model equipped with the Henyey-Greenstein phase function~\cite{Henyey1940DiffuseRI}. Evaluated directly against the calibrated metric depth $\mathbf{D}$, this explicit formulation yields distance-dependent visibility attenuation and localized light halos. The implementation details are provided in (Sec.~\ref{sec:fog_details} of the supplementary materials).

\textbf{Environment Harmonization.} To synthesize global ambient illumination for regions with sparse 3D geometry, we employ a neural forward renderer conditioned on an HDR environment map. The synthesized ambient radiance is linearly blended with the local light pass, effectively completing the deferred shading cycle. Implementation and fusion details are provided in (Sec.\ref{sec: environment_light_pass_editing} of the supplementary materials).

\subsection{VidRefiner}
\label{sec:videorefiner}

Dual-Pass Editing resolves physically consistent dynamics, yielding a deterministic baseline that necessitates terminal refinement to incorporate real-world sensor nuances. To prevent stochastic hallucinations from altering the resolved scene structure, we collapse the generative space toward the established physical manifold via two complementary constraints:

\textbf{Latent Initialization.} The rendered sequence serves as a comprehensive structural and spectral anchor, injecting low-frequency priors into the generative process. By perturbing VAE-encoded latents to a pivot timestep $t_s$, the reverse diffusion trajectory inherits the global layout, color distribution, and coarse lighting resolved in the physical simulation. This initialization restricts the generative process to high-frequency textural refinement, preventing unconstrained global synthesis while preserving the deterministic scene structure.

\textbf{Boundary Conditioning.} To complement the low-frequency priors, high-frequency spatial constraints are enforced via spatiotemporally coherent boundaries extracted from the rendered output. This integration utilizes a lightweight backbone~\cite{wan2025} pre-aligned for multi-channel conditioning, facilitating direct channel-wise concatenation without secondary fine-tuning. Unlike cross-attention mechanisms providing latent-level semantic guidance, this input-level formulation imposes an explicit spatial bias. The architectural choice of a lightweight model further restricts the synthesis of fine-grained textures to the resolved geometric limits, ensuring the structural integrity of edited elements remains invariant during photorealistic refinement.

The implementation details about the VidRefiner are provided in (Sec.~\ref{sec:postprocess} of the supplementary materials)

\section{Experiment}
\begin{table}[t]
    \centering

    \begin{tabular}{lccccc}
        \toprule  
         & S.A. & Night & Fog & Rain & Snow\\
        \midrule 
        Time (s) & 128.1 & 167.1 & 170.9 & 2.2 & 67.6 \\
        \bottomrule
    \end{tabular}

    \caption{Running time (seconds) per video for different tasks and weather conditions on a NVIDIA V100 GPU. S.A.: Semantic Annotation, which is computed once and shared by all four weather.}

    \label{tab:Running_time} 
    \vspace{-2em}

\end{table}

\subsection{Experimental Setting}
To validate our weather and time-of-day conversion method, we conduct experiments using PyTorch on NVIDIA GPUs (V100 for our method and most baselines; A100 for resource-intensive baselines like Cosmos-Transfer2.5~\cite{nvidia2025worldsimulationvideofoundation} and Ditto~\cite{bai2025ditto}). We evaluate on 120 scenes from the Waymo Open Dataset~\cite{10.1007/978-3-031-19818-2_4}, specifically using NOTR — a versatile subset of Waymo encompassing diverse driving scenarios, as surveyed in~\cite{emernerf}. The time consumption of the core component (G-buffer Dual-pass Editing) is reported in Tab.~\ref{tab:Running_time}. 

\subsection{Baselines}

\textbf{Baseline Selection.} Existing 3D-aware weather synthesis frameworks~\cite{dai2025rainygs, Li2023ClimateNeRF} are fundamentally constrained to static scenes, rendering them structurally incompatible with the highly dynamic environments of autonomous driving. Consequently, to evaluate temporal consistency and semantic fidelity in dynamic sequences, the framework is benchmarked against state-of-the-art video editing and foundation models: Video-P2P~\cite{liu2023videop2p}, Ditto~\cite{bai2025ditto}, Cosmos-Transfer2.5~\cite{nvidia2025worldsimulationvideofoundation}, and WAN-FUN 2.2~\cite{wan2025}. The comparative analysis is further extended to include domain-specific architectures: the inverse-rendering framework DiffusionRenderer~\cite{DiffusionRenderer} (constrained to translations conditioned on HDR environment maps) and the concurrent work WeatherEdit~\cite{weatheredit} (bounded to fog, snow, and rain synthesis).

\textbf{Evaluation Protocol.} The evaluation covers four primary weather and lighting conditions for autonomous driving: fog, midnight, rain, and snow. For baselines that require text inputs, prompts are constructed by directly concatenating the original scene description with the target weather condition. This standardized text input provides consistent guidance across all generative models, ensuring that performance differences are not caused by manual prompt engineering (Prompt details in Supplementary materials Sec.~\ref{sec:prompt}).

\textbf{Evaluation Metrics.} The synthesis fidelity and physical consistency of the generated sequences are systematically quantified across three dimensions. (1) \textit{Editing Instruction Adherence} utilizes the CLIP score~\cite{clipscore} to evaluate the precise execution of the targeted weather/time-of-day translation. (2) \textit{Structural Consistency} assesses geometric preservation through a bounding-box Intersection-over-Union (IoU) protocol. By comparing projected 2D ground-truth LiDAR boxes against extractions from a pre-trained monocular 3D detector~\cite{OVMono3D}, the relative IoU serves as a strict indicator of structural rigidity across all baselines (AABB projection formulations in Supp. Sec.~\ref{sec:more_quantitative}).  (3) \textit{Identity Stability} enforces the semantic invariance of foreground subjects. Computed via patch-level CLIP feature similarity before and after editing, this metric rigorously penalizes generative hallucinations while accommodating valid physical material alterations, such as snow accumulation or wet surface reflections.

\subsection{User Study}

A Two-Alternative Forced Choice (2AFC) study is conducted with 12 independent raters. To prevent scene selection bias, evaluation sequences are randomly sampled across diverse weather conditions and strictly standardized in resolution and duration ($512 \times 512$, 30 frames). The framework is benchmarked against the four baselines (10 paired comparisons per baseline). To negate cognitive bias, each trial enforces strict double-blind randomization for both presentation order and left-right layout. Evaluation is governed by two explicit criteria: (1) \textit{Spatial Fidelity}, assessing photorealism and editing instruction adherence; and (2) \textit{Temporal Coherence}, penalizing background flickering and evaluating the motion continuity of dynamic weather. Final win rates are aggregated from 1,440 independent responses.

\subsection{Results}
\textbf{Qualitative Results.} We present qualitative results of our editing framework in Fig.~\ref{fig:teaser_results}. As illustrated, given an autonomous driving video, our approach enables precise control over key scene attributes—including shadows, lighting, and geometry—to facilitate conversions between different weather conditions and times of day. 

\begin{figure*}[t]
  \centering
    \includegraphics[width=1.0\linewidth]{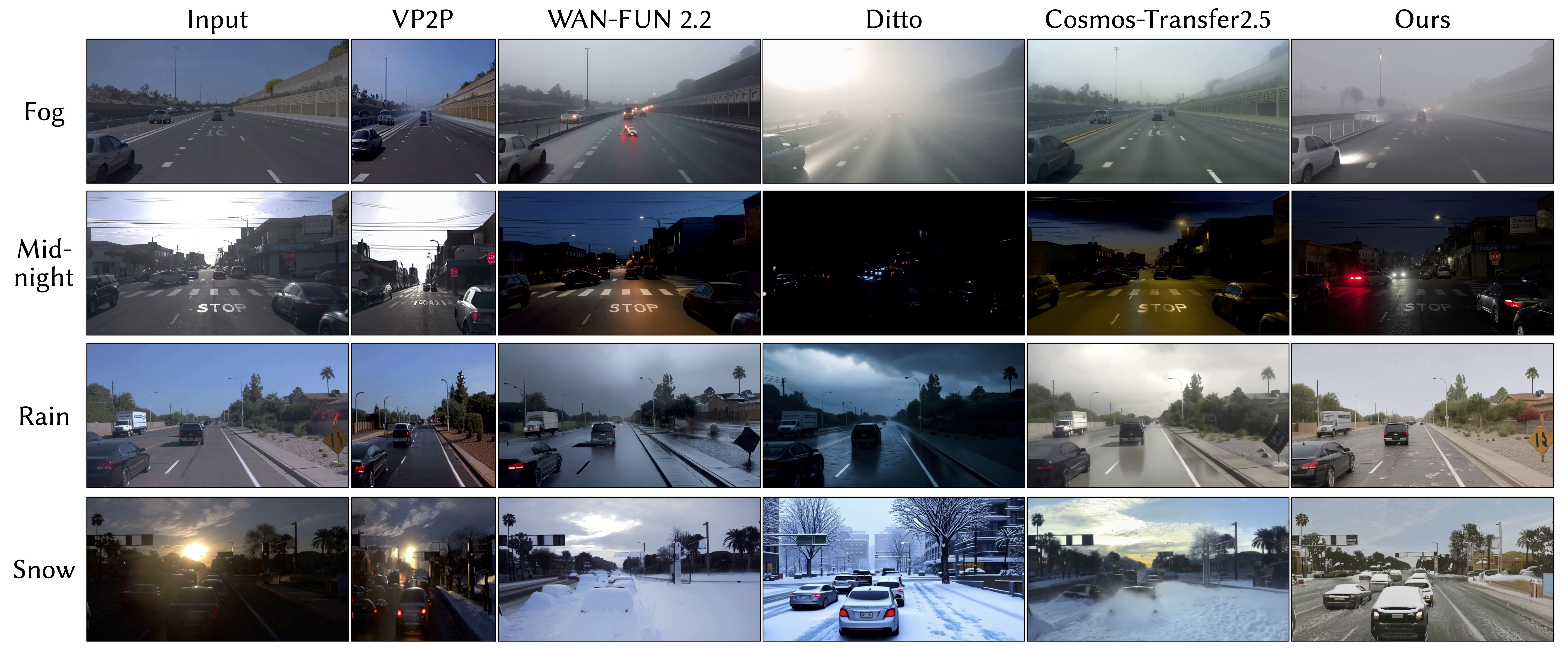}
    \caption{\textbf{Qualitative Comparisons of AutoWeather4D on Waymo Weather/ Time-of-day Conversions: Validating Physically Plausible and Fine-Grained Control for Autonomous Driving.}}
    \label{fig:qualitative}
\end{figure*}

\begin{figure*}[t]
  \centering
    \includegraphics[width=1.0\linewidth]{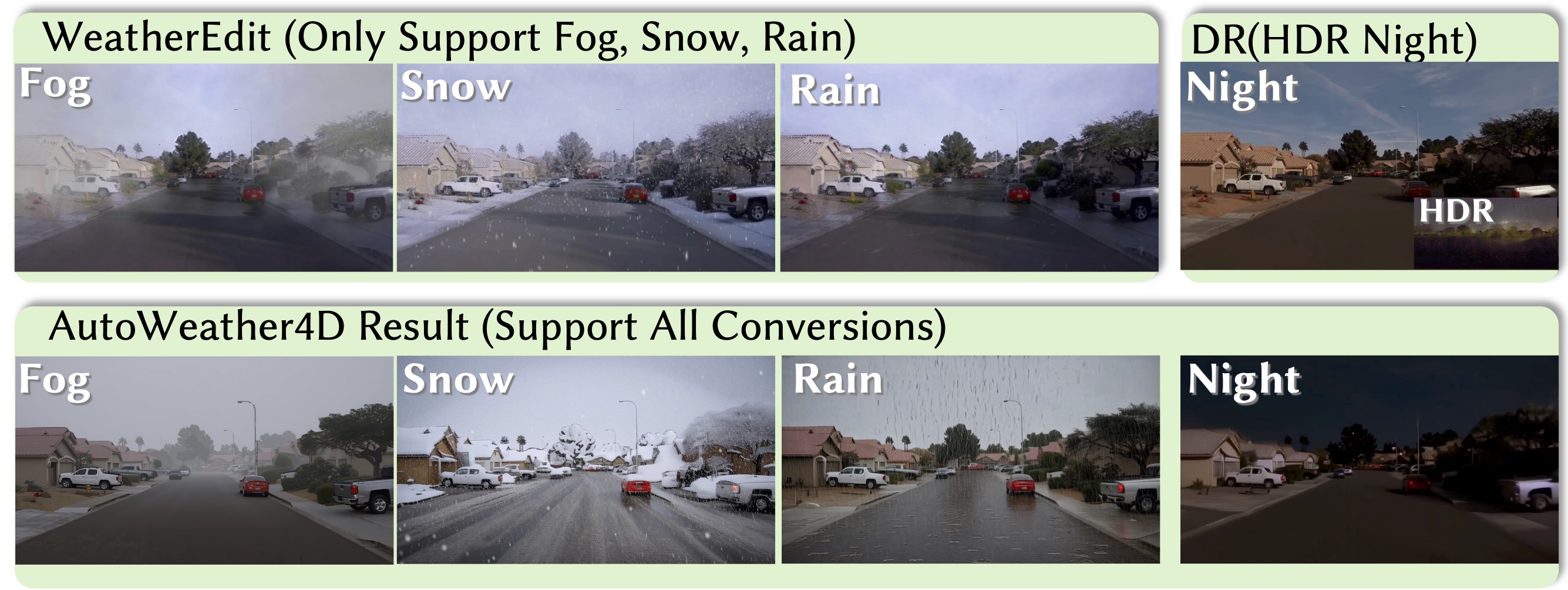}
    \caption{\textbf{Qualitative Comparisons with domain-specific architectures: Validating spatially anchoring weather effects.} DR: DiffusionRenderer~\cite{DiffusionRenderer}.}
    \label{fig:weatheredit}
    \vspace{-1em}
\end{figure*}

\textbf{Comparisons Results.} We provide qualitative examples in Fig.~\ref{fig:qualitative} to demonstrate the effectiveness of our method. Video-P2P struggles to follow instructions, while Ditto introduces background structural artifacts and generative hallucinations (e.g., extraneous architectural elements in row 4). In contrast to WAN-FUN and Cosmos-Transfer, our approach enforces physical plausibility and fine-grained spatial control. Specifically, row 1 illustrates how our explicit parameterization allows for localized manipulation of individual light sources and fog density, effectively resolving the ambiguity of pure text-conditioning. Furthermore, as shown in rows 2 and 3, the baselines fail to disentangle source lighting, resulting in biased road brightness or incompatible hard shadows. Our method, however, correctly models illumination transport. Finally, row 4 validates that our pipeline strictly preserves foreground geometry. These structural and photometric consistencies are directly attributed to our explicit illumination decomposition and parameterized light control modules.

\textbf{Comparison with Domain-Specific Architectures.} As illustrated in Fig.~\ref{fig:weatheredit}, WeatherEdit retains hard shadows under highly scattering conditions (e.g., fog, rain, snow). In contrast, the decoupled G-buffer representation mitigates the shape-radiance ambiguity~\cite{zhang2020nerf++} by anchoring the editing process in explicit geometry.  This structural prior further facilitates consistent diffuse shading and enables localized weather interactions, such as snow accumulation and surface ripples. Furthermore, compared to DiffusionRenderer~\cite{DiffusionRenderer}, the 3D-aware formulation leverages spatial priors to isolate the sky region. This depth-based separation avoids unintended sky relighting and supports localized illumination driven by ego-vehicle headlights. Extended results are provided in (Supp. Sec.~\ref{sec:more_qualitative}) and the supplementary videos.

\begin{table}[htbp]
  \centering

  \begin{tabularx}{\linewidth}{@{}l *{4}{>{\Centering}X}@{}}
    \toprule
    Model & CLIP Score (↑) & Vehicle 3D Detection IoU (↑) & Vehicle CLIP cosine similarity (↑) & Human Evaluation (↑) \\
    \midrule
    Video-P2P & 0.2448 & - & - & 0 \\
    Ditto & 0.2532 & 0.805 & 0.769 & 0.425 \\
    Cosmos-Transfer2.5 & 0.2558 & 0.913 & 0.837 & 0.580 \\
    WAN-FUN 2.2 & 0.2577 & 0.888 & 0.794 & 0.668 \\
    Ours & \textcolor{red}{\textbf{0.2586}} & \textcolor{red}{\textbf{0.915}} & \textcolor{red}{\textbf{0.871}} & \textcolor{red}{\textbf{0.826}} \\
    \bottomrule
  \end{tabularx}

    \caption{\textbf{Quantitative evaluation of weather and time-of-day conversions on the Waymo dataset}. All results are derived from 120 source videos, each converted into four common adverse conditions (rain, snow, fog, night) with 57 frames per video. This results in a total of 27,360 frames evaluated. ``-" indicates omitted due to frame cropping. All metrics are averaged across all frames, where \textcolor{red}{\textbf{red}} indicates the best performance. }
  \label{tab:eval_general_uniform}

  \vspace{-2em}
\end{table}
\textbf{Quantitative Result Analysis.} As shown in Tab.~\ref{tab:eval_general_uniform}, AutoWeather4D achieves performance comparable to existing baselines. While numerical results are on par with massive data-driven models (e.g., Cosmos-Transfer2.5), they demonstrate that our framework maintains high generation fidelity. Crucially, our method complements the implicit generation process by introducing explicit physical control, offering a deterministic alternative for fine-grained editing. Extended comparative evaluations on general generative performance—encompassing FVD and distribution-level metrics distinct from core editing fidelity—are deferred to (Supplementary Materials Sec.~\ref{sec:more_quantitative}).


\subsection{Ablation Study}

Comprehensive architectural ablations—encompassing isolated modules, combinatorial configurations, and granular physical parameterizations (rain, snow, local illumination, and VidRefiner conditioning strength)—are strictly deferred to the Supplementary Materials (Sec.~\ref{sec: ablation_studies}). The subsequent analysis explicitly isolates the structural necessity of the continuous 4D reconstruction.

\begin{figure}[t]
  \centering
    \includegraphics[width=0.75\linewidth]{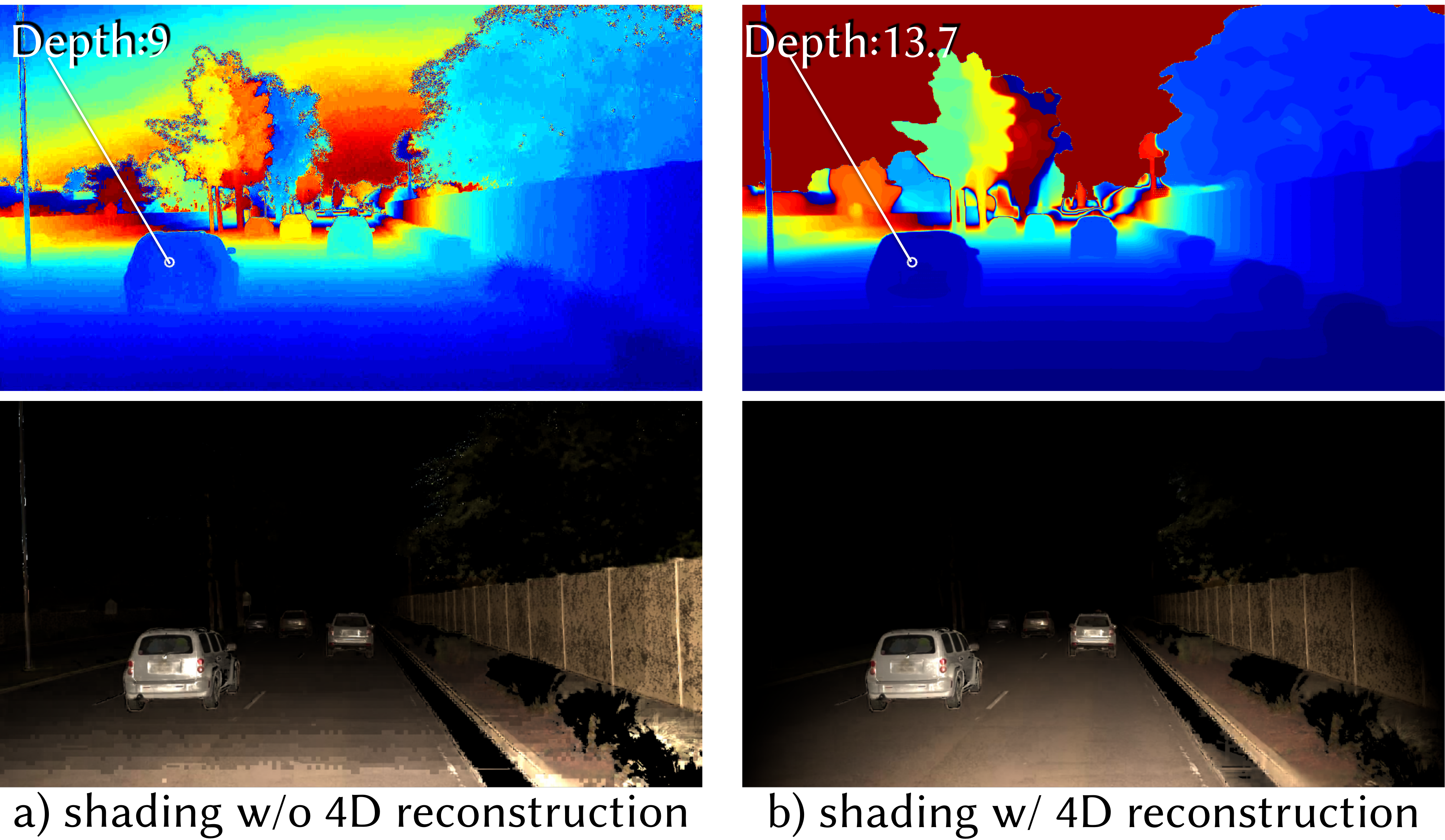}
    \caption{\textbf{Ablation of 4D reconstruction.} (a) Integer-quantized depth priors~\cite{DiffusionRenderer} induce severe spatial discretization and aliasing during local relighting. (b) The deployed feed-forward 4D reconstruction establishes a continuous floating-point manifold, enforcing smooth, artifact-free illumination gradients.}
    \label{fig:reconstruction_ablation_study}
\end{figure}

\textbf{Effect of 4D reconstruction.} Distance-based light attenuation dictates a strict mathematical requirement for continuous spatial gradients. Utilizing standard inverse rendering alone extracts integer-quantized depth maps. Under explicit local illumination, this spatial discretization inherently provokes abrupt discontinuities in the light attenuation function, manifesting as severe jagged aliasing across the rendered surfaces (Fig.~\ref{fig:reconstruction_ablation_study}a). By integrating the feed-forward 4D reconstruction backbone, the pipeline recovers a continuous, floating-point geometric manifold. This non-discrete structural prior deterministically resolves spatial step artifacts, guaranteeing natural and continuous illumination gradients during dynamic relighting (Fig.~\ref{fig:reconstruction_ablation_study}b).

\subsection{Applications}

\textbf{Efficacy in Perception Data Augmentation.} Following established weather synthesis protocols~\cite{weatheredit}, the utility of AutoWeather4D is assessed for downstream perception data augmentation. The synthesized sequences serve as supplementary training data to address domain gaps in semantic segmentation. Source-domain semantic annotations are directly transferred to the synthesized adverse videos. Subsequently, the HRDA segmentation model~\cite{hrda} is fine-tuned on 6,480 augmented frames for 20k iterations. The segmentation performance is evaluated via mIoU and mAcc using the Cityscapes~\cite{cityscapes} taxonomy across two standard adverse-condition datasets: ACDC~\cite{acdc} (snow, rain, fog, night) and Dark Zurich~\cite{darkzurich} (day, night).

\begin{table}[htbp]

\scriptsize
  \centering
  \begin{tabular}{lcccc}
    \toprule
    Setting & ACDC mIoU(↑) & ACDC mAcc(↑) & DarkZurich mIoU(↑) & DarkZurich mAcc(↑) \\
    \midrule 
    w/o augmentation & 49.20 & 60.72 & 23.92 & 38.29 \\
    w/ cosmos & 49.66\ (+0.93\%) & 62.31\ (+2.62\%) & 23.93\ (+0.04\%) & 39.52\ (+3.21\%) \\
    w/ ours & \textbf{\textcolor{red}{49.81\ (+1.24\%)}} & \textbf{\textcolor{red}{62.52\ (+2.96\%)}} & \textbf{\textcolor{red}{24.09\ (+0.71\%)}} & \textbf{\textcolor{red}{39.73\ (+3.76\%)}} \\
    \bottomrule
  \end{tabular}

\caption{\textbf{AutoWeather4D for data augmentation in adverse weather semantic segmentation.}}
  \label{tab:downstream}

  \vspace{-2em}
\end{table}

Tab.~\ref{tab:downstream} reports the quantitative impact of the synthesized sequences on downstream perception robustness. While absolute mIoU increments on ACDC and Dark Zurich remain marginal, this reflects the inherent saturation of zero-shot cross-domain transfer within a highly optimized baseline (HRDA). Consequently, this proxy evaluation is formulated strictly to validate the geometric fidelity of the synthesized data, rather than to advance the segmentation state-of-the-art. Under severe atmospheric shifts, standard generative baselines exhibit structural degradation, resulting in negligible transfer benefits (e.g., Cosmos yielding a +0.04\% mIoU increment on Dark Zurich). Conversely, the consistent gains provided by AutoWeather4D demonstrate that explicitly anchored synthesis reliably preserves the underlying scene geometry required for robust downstream training. 

\begin{figure*}[t]
  \centering
    \includegraphics[width=0.96\linewidth]{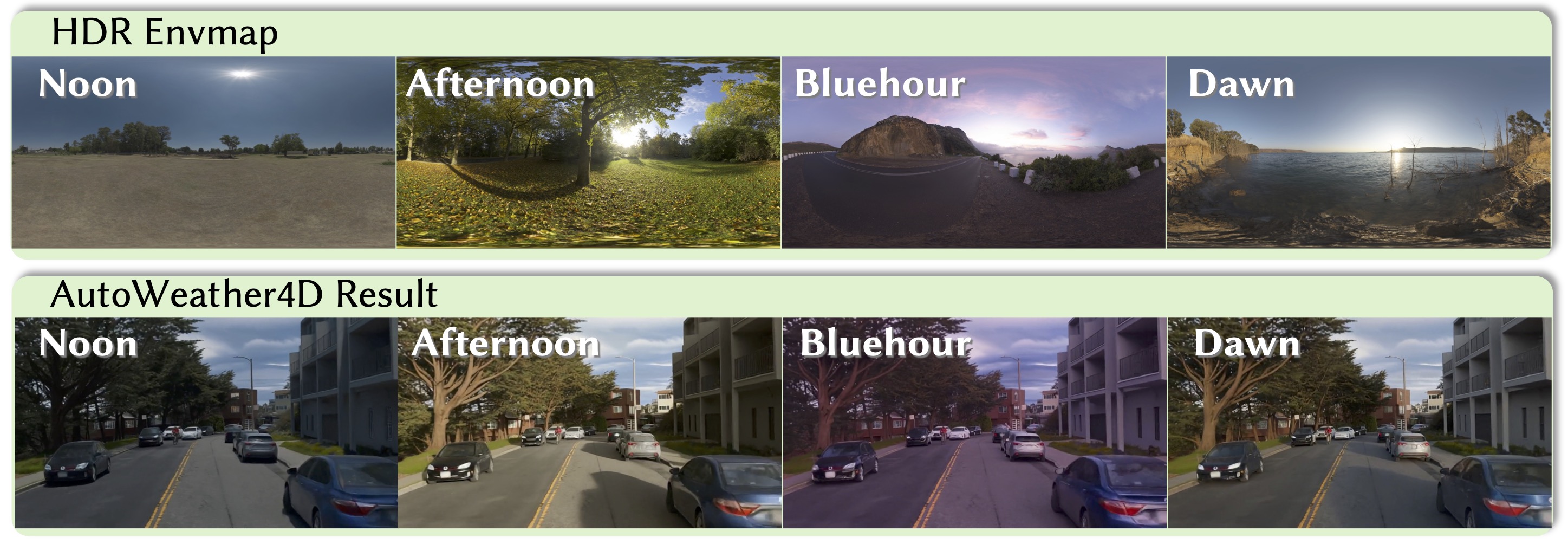}
    \caption{\textbf{Global Illumination Editing driven by HDR environment map}.}
    \label{fig:time-of-day}
\end{figure*}

\textbf{Illumination Control.} The decoupled representation enables HDR-driven illumination. Parameters such as sun altitude and ambient color are adjustable without changing scene geometry (Fig~\ref{fig:time-of-day}). The G-buffer ensures consistent shading and light transport across different environment maps.

\begin{figure*}[t]
  \centering
    \includegraphics[width=0.96\linewidth]{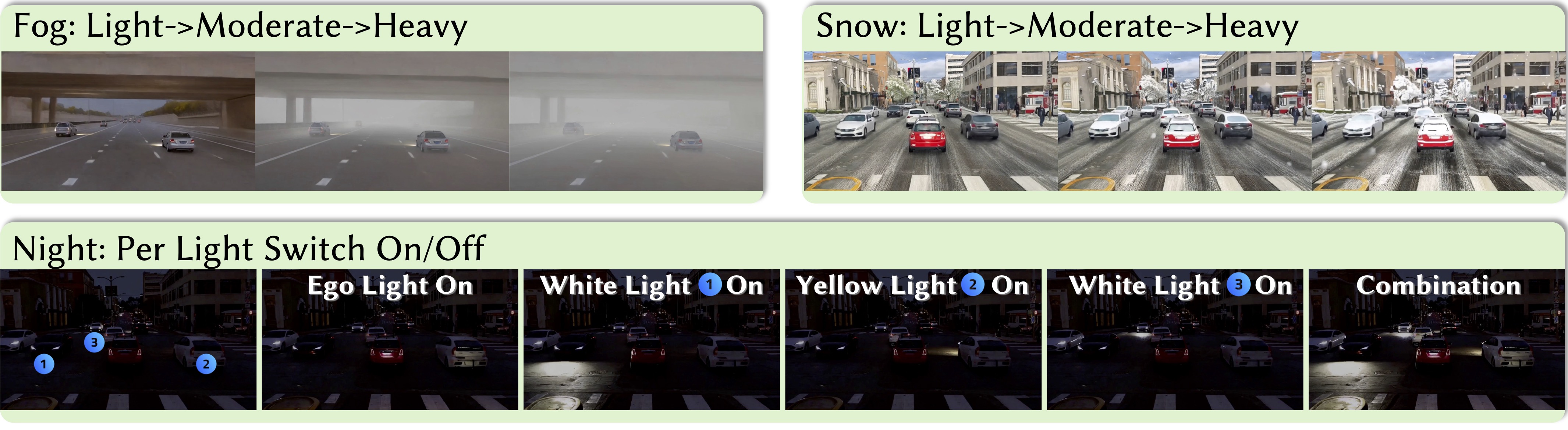}
    \caption{\textbf{Explicit Parameterized Control for Physically Consistent Weather and Time-of-Day Video Editing}.}
    \label{fig:variations}
    \vspace{-2em}
\end{figure*}

Furthermore, the architectural advantage of AutoWeather4D extends beyond static data augmentation. As demonstrated in Fig.~\ref{fig:variations}, our explicit parameterized control (e.g., continuously scaling fog density or toggling local lights) enables the targeted synthesis of specific challenging scenarios where downstream algorithms commonly fail. While comprehensive downstream diagnostic benchmarking is left for future work, this fine-grained, deterministic controllability inherently provides the foundation for generating continuous perturbation sequences. This establishes a highly controllable capability for future system-level robustness evaluation, providing a complementary deterministic approach to purely data-driven generative baselines, avoiding the need for excessively large-scale specific data collection.

\vspace{-1em}
\section{Conclusion}
\label{sec:conclusion}
\vspace{-0.5em}

To mitigate data scarcity, AutoWeather4D aligns deterministic graphics with video diffusion to transform real-world footage into adverse scenarios. By anchoring the generative process to explicit G-buffer priors, we transition from synthesis with entangled geometry and illumination toward physically grounded, parametric simulation. This framework serves as a complementary data source, offering potential for future diagnostics of perception failure modes in adverse conditions.

\textbf{Limitations.} While our explicit-implicit bridging achieves physically grounded synthesis for primary weather conditions, capturing extreme long-tail dynamic interactions (e.g., complex fluid dynamics of vehicle splash) remains challenging for decoupled pipelines. Future work will explore integrating localized generative priors specifically for these unstructured phenomena, complementing our deterministic structural anchors. Additionally, balancing severe environmental perturbations (e.g., heavy fog occlusion) with the structural retention of distant background elements necessitates careful calibration. While our boundary conditioning effectively anchors foreground geometries, future iterations could incorporate semantic-aware attenuation masks to dynamically modulate the diffusion intensity across critical autonomous driving regions of interest.

%
%
\bibliographystyle{splncs04}
\bibliography{main}
\newpage
\begin{center}
{\Large \textbf{Supplementary Materials for AutoWeather4D}}
\end{center}
\setcounter{section}{5}
\setcounter{figure}{6}
\setcounter{table}{4}

This document provides implementation details, extended quantitative evaluations, and ablation studies to support the reproducibility of AutoWeather4D. The structure is organized as follows:

\begin{itemize}
    \item Sec.~\ref{sec:4d_reconstruction_details}: Details of feed-forward G-buffer extraction, encompassing metric calibration, alignment, and sky-aware material extraction.
    \item Sec.~\ref{sec:snow_details}: Physical modeling and implementation for snow synthesis via G-buffer dual-pass editing.
    \item Sec.~\ref{sec:rain_details}: Physical modeling and implementation for rain synthesis via G-buffer dual-pass editing.
    \item Sec.~\ref{sec: semantic annotations}: Pipeline and configurations for dense semantic annotations.
    \item Sec.~\ref{sec: night_details}: Implementation specifics for nocturnal local relighting via G-buffer dual-pass editing.
    \item Sec.~\ref{sec:fog_details}: Implementation specifics for volumetric fog synthesis via G-buffer dual-pass editing.
    \item Sec.~\ref{sec: environment_light_pass_editing}: Procedures for environment harmonization.
    \item Sec.~\ref{sec:prompt}: Strategies and templates for text prompt design.
    \item Sec.~\ref{sec:postprocess}: Architecture and parameter configurations for the VidRefiner module.
    \item Sec.~\ref{sec:more_quantitative}: Extended quantitative evaluations and metrics.
    \item Sec.~\ref{sec: ablation_studies}: Comprehensive ablation studies on architectural components.
    \item Sec.~\ref{sec:more_qualitative}: Extensive qualitative comparisons and visualizations.
\end{itemize}

\section{Feed-Forward G-Buffer Extraction}
\label{sec:4d_reconstruction_details}
To ensure consistent light effect parameters across different sequences, we calibrate the reconstructed metric depth using real-world physical scales. We introduce a calibration step using LiDAR-captured point clouds. For calibration, we sample $N(N=1000 \text{ empirically})$ non-sky and non-occluded points from the LiDAR data via RANSAC~\cite{ransac}, match each sampled point to the corresponding depth value from the 4D reconstruction result, and minimize the mean squared error between the two depth sets to solve for scale $s$ and bias $b$. This optimization follows the loss function:

\begin{equation}
\mathcal{L} = \frac{1}{N}\sum_{i=1}^N (s \cdot d_{\text{4D},i} + b - d_{\text{LiDAR},i})^2.    
\end{equation}

Here, $d_{\text{4D},i}$ and $d_{\text{LiDAR},i}$ represent the reconstructed depth and LiDAR depth of the i-th point, respectively. This calibration ensures the 4D geometry adheres to real-world dimensions, a key requirement for physically plausible editing tasks such as maintaining consistent object size across frames.

Feed-forward 4D reconstruction often struggles with sky depth estimation, and incorrect depth values can cause the sky to be erroneously illuminated by light sources. This challenge arises from the sky’s limited texture and large depth variance, which typically result in fragmented or unrealistic depth values. To address this issue, we explicitly segment the sky mask for each frame using Grounded-SAM~\cite{ren2024grounded} with the text prompt ``sky''. After segmentation, we set the sky depth to the 99th percentile of the non-sky depth distribution across the entire sequence. This approach avoids extreme depth values that would disrupt lighting calculations while preserving visual consistency with the scene’s real depth range.

\textbf{Monocular Fallback Calibration via Camera Height Prior}. In strictly monocular configurations where sparse LiDAR point clouds are unavailable, we recover the absolute metric scale by leveraging the known camera height $H_{cam}$ as a geometric prior. This is achieved by anchoring the relative depth of the road surface to the physical camera height.

First, we utilize the dense semantic annotations (as described in Sec. 9) to isolate the road mask $M_{road}$. For each pixel $i \in M_{road}$ with coordinates $(u_i, v_i)$, we project it into the unscaled 3D space to obtain its relative 3D coordinate $\mathbf{P}_{rel,i}$ using the predicted relative depth $d_{4D,i}$ and the camera intrinsic matrix $K$:

\begin{equation}
\mathbf{P}_{rel,i} = d_{4D,i} K^{-1} \begin{bmatrix} u_i \\ v_i \\ 1 \end{bmatrix}
\end{equation}

Next, we apply RANSAC to fit a ground plane to these unscaled 3D road points, estimating the relative ground normal vector $\mathbf{n}$ (where $||\mathbf{n}|| = 1$). The relative camera height $h_{rel}$ in the unscaled 3D space is derived as the orthogonal distance from the camera origin to the fitted ground plane:

\begin{equation}
h_{rel} = \frac{1}{|M_{road}^{inliers}|} \sum_{i \in M_{road}^{inliers}} | \mathbf{n}^T \mathbf{P}_{rel,i} |    
\end{equation}

Finally, we deterministically resolve the global scale factor $s$ by taking the ratio of the known physical camera height $H_{cam}$ (e.g., typically 1.5m to 2.0m for autonomous driving vehicles) to the estimated relative height $h_{rel}$. In this fallback configuration, we assume the bias offset $b \approx 0$ to avoid underdetermined equation solving:

\begin{equation}
s = \frac{H_{cam}}{h_{rel}}    
\end{equation}

The absolute metric depth is then deterministically recovered as $d_{metric} = s \cdot d_{4D}$. This fallback mechanism ensures that our framework remains highly adaptable to purely monocular video streams while enforcing the strict metric alignment required for physically valid illumination transport.

\section{Snow Synthesis via G-Buffer Dual-Pass Editing}
\label{sec:snow_details}
\subsection{Metaball-based surface snow}
SPH Poly6 kernel as our metaball implicit function shows in Equation~\ref{eq:poly6}, which provides smooth blending and efficient gradient computation:

\begin{equation}
\label{eq:poly6}
W(r,\rho)=
\begin{cases}
\dfrac{315}{64\pi \rho^{9}}\,(\rho^{2}-r^{2})^{3}, & 0\le r < \rho,\\[6pt]
\hfill0,\hfill & r\ge \rho
\end{cases}
\end{equation}

where $r$ is the distance from the evaluation point to the metaball center, and $\rho$ is the support radius. We set $\rho = 0.1\,\text{m}$ (twice the particle radius of $0.05\,\text{m}$) to facilitate smooth topological merging of the snow accumulation. The radial derivative for gradient computation is:

\begin{equation}
\label{eq:dpoly6}
\frac{dW}{dr}(r,\rho) = -\dfrac{945}{32\pi \rho^{9}} \, r \, (\rho^{2}-r^{2})^{2}, \quad 0 < r < \rho.
\end{equation}

This kernel is applied in a cascaded manner with decaying amplitudes to simulate multi-scale snow buildup. For each surface point, the snow height field is computed as:

\begin{equation}
H_{\text{snow}}(\mathbf{x}) = \sum_{l=0}^{L-1} \lambda^l \sum_{i \in \mathcal{N}_k(\mathbf{x})} a_i \cdot W(|\mathbf{x} - \mathbf{c}_i|, \rho_l)
\end{equation}

where $\mathcal{N}_k(\mathbf{x})$ denotes the $k$-nearest metaballs (we set $k=16$), In our implementation, we use $L=3$ cascade levels. The amplitude decay factor is set to $\lambda=0.7$, The density weights $a_i$ are randomly jittered in [0.8,1.2] to introduce natural variation. And $\rho_l = \rho_0 / \xi^l$ are cascaded radii with a base support radius $\rho_0=0.5$ and a scaling factor $\xi=1.5$. Surface normals are perturbed using the tangent-space gradient, clamped to prevent unrealistic slopes.

\subsection{Material blending}
Material properties are modified via a soft blending mechanism using a weighted sigmoid function:

\begin{equation}
\label{eq:sigmoid_blend}
\sigma(x; w, \tau_{\text{bias}}) = \frac{1}{1 + \exp(-w (x - \tau_{\text{bias}}))},
\end{equation}

where $x$ is the computed snow height $H_{\text{snow}}$, $w$ is the blend weight ($w=0.8$), and $\tau_{\text{bias}}$ is the threshold bias ($\tau_{\text{bias}}=0.03$). Coverage is gamma-corrected ($\gamma=0.9$) and thresholded for hard edges if desired. Albedo is lerped towards a uniform snow value of 1.0, roughness is adjusted to 0.6 to simulate diffuse snow, and metallic is reduced to zero.  Optional displacement along the original normal adds geometric detail.

\subsection{Wet ground implementation}
Wet ground simulation follows the physically-based wet surfaces model. It darkens albedo based on porosity and water intensity using:

\begin{equation}
\label{eq:wet_albedo}
A_{\text{wet}} = A_{\text{dry}} \cdot (1 - p) + A_{\text{water}} \cdot p \cdot e^{-\tau_{\text{opt}} / \mu},
\end{equation}

where $A_{\text{dry}}$ is the original base color texture, we set porosity $p=0.8$. $A_{\text{water}}$ is water albedo, We assume $A_{\text{water}} \approx 0.02$ due to high absorption. $\tau_{\text{opt}}$ is optical depth (approximate to 0), and $\mu = \cos\theta$ accounts for view angle. Roughness is reduced to mimic water sheen via linear interpolation:

\begin{equation}
\label{eq:wet_roughness}
r_{\text{wet}} = r_{\text{dry}} \cdot (1 - i) + r_{\text{water}} \cdot i,
\end{equation}

where $i$ is the wetness intensity (set to 0.5) and $r_{\text{water}}$ is near 0 for smooth water ( set to 0.1 in our implement). This is applied selectively to non-snow-covered ground areas.

\subsection{Particle-based falling snow}
For particle-based falling snow, we use below formula to simulate:

\begin{equation}
\label{eq:particle_motion}
\mathbf{p}_{t+1} = \mathbf{p}_t + (\mathbf{v}_{\text{gravity}} + \mathbf{v}_{\text{wind}}) \cdot \Delta t
\end{equation}

where $\mathbf{p}_t \in \mathbb{R}^3$ is the particle world-space position at discrete time step $t$, and $\Delta t$ is the simulation time step. We simulate $6,000$ particles within a view frustum-aligned bounding box to ensure coverage. The downward drift velocity is set to $\mathbf{v}_{\text{gravity}} = [0, -2.0, 0]^\top\,\text{m/s}$, and the wind advection velocity is set to $\mathbf{v}_{\text{wind}}=[0.3, 0, 0.1]^\top\,\text{m/s}$ to introduce lateral movement.

\section{Rain Synthesis via G-Buffer Dual-Pass Editing}
\label{sec:rain_details}

We provide additional implementation details for our rain rendering pipeline. All parameters were empirically tuned on urban driving sequences.

\subsection{Geometry-Anchored Puddle Modeling via World-Space FBM}

While our feed-forward 4D reconstruction successfully extracts the macro-topology of the dynamic scene, millimeter-level micro-geometry (e.g., subtle road depressions and potholes) remains inherently unobservable from standard monocular driving videos. To bridge this physical resolution gap and strictly adhere to our surface-anchored interaction paradigm, puddle boundaries are procedurally synthesized by projecting Fractional Brownian Motion (FBM)~\cite{fbm} into the explicit 3D world space, rather than applying it as a 2D screen-space overlay.

Specifically, for each pixel $i$ classified as ``road'' with camera-space 3D coordinates $\mathbf{P}_{c,i} = [X_c, Y_c, Z_c]^T$ derived from the metric depth $D$, we first transform it into a temporally coherent world coordinate system $\mathbf{P}_{w,i} = [X_w, Y_w, Z_w]^T$ using the estimated camera pose. The FBM noise is exclusively evaluated along the lateral planar coordinates $(X_w, Z_w)$ of the road surface:

\begin{equation}
\mathcal{N}_{puddle}(X_w, Z_w) = \sum_{o=1}^{O} \frac{1}{2^o} \text{Noise}(2^o \cdot [X_w, Z_w]^T)
\end{equation}

where $O=3$ is the number of octaves, with persistence $\alpha=0.5$ and lacunarity $\lambda=2.0$. The base noise is value noise evaluated at a physical scale of $0.05~m^{-1}$.

By sampling the noise directly on the explicit 3D manifold, the generated puddles are rigorously anchored to the underlying scene geometry. This world-space parameterization guarantees that the procedural water pools intrinsically exhibit correct perspective foreshortening, physical occlusion, and strict temporal consistency across dynamic camera movements.

Finally, we apply power redistribution with a unit exponent followed by cascaded smoothstep operations at physical thresholds $(0.0, 0.7)$ and $(0.2, 1.0)$  to extract the binary and transitional puddle masks $M_{puddle}$, integrating this micro-geometric hallucination with our deterministic G-buffer mechanics.

\subsection{Precipitation Dynamics}

We simulate $10^4$ raindrops with diameters uniformly sampled from $[0.5, 6.0]$ mm. Terminal velocities follow the Gunn-Kinzer model as described in the main paper, with horizontal wind modeled as a base velocity of $(0.1, 0, 0)$ m/s plus Gaussian perturbations $\mathcal{N}(0, 0.5^2)$. Drops initialize at heights uniformly distributed in $[0, 51]$ m and reset upon collision or boundary exit.

Each raindrop is rendered as an uneven capsule with tail radius $r_t = r_h / 0.7$ to simulate motion blur. The streak length is computed as $0.8\cdot \Delta t \cdot v$ where $\Delta t$ is the frame interval and $v$ is the drop velocity. This asymmetric geometry creates more realistic visual streaks compared to uniform capsules.

For rendering, drops are projected into screen space, and signed distance fields (SDFs) are computed per pixel. The SDF for an uneven capsule is:
\begin{equation}
\label{eq:sdf_capsule}
\text{sdf}(\mathbf{p}) = d_{\text{axis}}(\mathbf{p}) - r_{\text{interp}}(\mathbf{p}),
\end{equation}
where $d_{\text{axis}}$ computes the distance to the capsule's central axis between head and tail positions, and $r_{\text{interp}}$ interpolates between head radius $r_h$ and tail radius $r_t = r_h/\gamma$ with taper factor $\gamma \approx 0.7$. Negative SDF values trigger G-buffer updates including alpha blending for translucency and depth biasing for proper occlusion.

\subsection{Surface Perturbation Effects}

Ripple generation employs a grid-based procedural approach with 32-pixel cells. Within each cell, we generate expanding ring waves at 31.0 rad/m frequency with radial falloff using smoothstep windowing in $[-0.6, 0.0]$. The ripple intensity oscillates between $[0.01, 0.15]$ based on temporal modulation, and the resulting normal perturbations blend with base normals using strength factor 0.9 in puddle regions.

\subsection{Material Response Parameters}

Our G-buffer modifications vary by surface type to achieve realistic wet appearance:

\textbf{Atmospheric Effects.} Sky pixels are tinted toward overcast conditions using color $(0.55, 0.60, 0.70)$ at 60\% strength with 70\% desaturation, simulating the diffuse lighting typical of rainy weather.

\textbf{Surface Wetness.} Ground surfaces in puddle regions have roughness reduced to 0.0 with a global wetness factor of 0.2 applied to non-puddle areas. Ripple highlights add subtle white tint $(0.92, 0.96, 1.00)$ at wave crests to enhance specular response.

\textbf{Raindrop Rendering.} Individual drops are alpha-blended at 40\% opacity with depth bias of $10^{-4}$ m for proper occlusion handling. Batch processing with chunk size 256 enables efficient GPU utilization.

\section{Dense Semantic Annotations}
\label{sec: semantic annotations}
We establish dense semantic annotations through a detect-segment-propagate pipeline that provides structural priors for downstream weather synthesis. These semantic tracks anchor spatial constraints that are critical for illumination and precipitation rendering: street-light instances localize emissive sources for headlamp estimation, road segments confine accumulation regions, and vehicle masks identify dynamic occluders requiring headlight relighting.

\textbf{Initialization via zero-shot detection.} 
At uniformly sampled keyframes, OWL-ViT~\cite{10.1007/978-3-031-20080-9_42} produces class-aligned proposals 
for ``street light'', ``road'', and ``car'' categories. To suppress ambiguity, 
we retain only the maximum-area road region per frame while filtering 
low-confidence detections. Each surviving box seeds SAM2's  \cite{ravi2024sam2} image predictor to obtain precise instance masks. 

\textbf{Bidirectional mask propagation.} 
The SAM2 video predictor extends keyframe masks across time via optical 
flow tracking, operating both forward and backward from each anchor frame. 
This strategy mitigates temporal flicker and maintains object identity 
across occlusions, yielding dense per-frame annotations for downstream 
depth analysis.
 This bidirectional propagation mitigates flicker, preserves object identities, and delivers dense per-frame annotations for street lights, roads, and vehicles, forming the foundation for the subsequent analysis modules.

\textbf{3D Reprojection of Instance Masks and Light Source Estimation}
To rigorously estimate the 3D spatial coordinates of local illuminants (e.g., street lights), we explicitly formulate a geometry-aware extraction pipeline utilizing the per-frame semantic masks and our reconstructed metric depth.

Step 1: 3D Reprojection and Point Cloud Aggregation.

Given the street light instance mask $M_t$ at frame $t$, for each valid pixel with 2D coordinates $\mathbf{u}_i = [u_i, v_i]^T \in M_t$, we utilize the calibrated metric depth $d_{i,t}$ and the camera intrinsic matrix $K$ to reproject the pixel into the global 3D space. The 3D coordinate $\mathbf{X}_{i,t}$ is computed as:

\begin{equation}
\mathbf{X}_{i,t} = d_{i,t} K^{-1} \begin{bmatrix} u_i \\ v_i \\ 1 \end{bmatrix}    
\end{equation}

By transforming all frames into a unified world coordinate system using the estimated camera poses, we aggregate a global raw point cloud for all potential street light structures, denoted as $\mathbb{P} = \{ \mathbf{X}_{i,t} \mid \forall t, \forall \mathbf{u}_i \in M_t \}$.

Step 2: Instance Grouping via Disjoint-Set Union (DSU).

To disentangle the unorganized global point cloud $\mathbb{P}$ into distinct static light instances, we construct a spatial undirected graph $\mathcal{G} = (\mathbb{P}, \mathcal{E})$. An edge exists between two points $\mathbf{X}_a$ and $\mathbf{X}_b$ if their Euclidean distance is strictly less than a spatial threshold $\tau_{dist}$ (empirically set to $0.5m$ to account for reconstruction noise):

\begin{equation}
\mathcal{E} = \{ (\mathbf{X}_a, \mathbf{X}_b) \mid ||\mathbf{X}_a - \mathbf{X}_b||_2 < \tau_{dist} \}
\end{equation}

We then employ the Disjoint-Set Union (DSU) algorithm to efficiently compute the connected components of $\mathcal{G}$. This process merges fragmented per-frame observations into spatially coherent, distinct 3D instance clusters $\mathcal{C}_k$, where $k \in \{1, 2, \dots, N\}$ represents the index of each uniquely identified street light.

Step 3: Light-Emitting Point Localization.

The physical light-emitting bulb is typically located at the apex of the street light structure, overhanging the road. To reliably estimate this emission center $\mathbf{p}_k$ for cluster $\mathcal{C}_k$ while maintaining robustness against geometric outliers, we isolate the subset $\mathcal{C}_{k, top}$ containing the top $5\%$ of points evaluated along the global upward axis $\mathbf{n}_{up}$ (orthogonal to the road plane $\mathbf{n}$ estimated in Sec. 6). The 3D position of the illuminant is analytically resolved as the spatial centroid of this apex subset:

\begin{equation}
\mathbf{p}_k = \frac{1}{|\mathcal{C}_{k, top}|} \sum_{\mathbf{X} \in \mathcal{C}_{k, top}} \mathbf{X}
\end{equation}

Finally, exploiting the spatial relationship between the street lights and the road, the spotlight direction vector $\mathbf{d}_k$ is deterministically parameterized to point downwards toward the road surface, ensuring physically plausible angular attenuation $S_j(x)$ during the subsequent Light Pass rendering (as formulated in Eq.~\ref{eq:light_pass}).

\section{Nocturnal Local Relighting via G-Buffer Dual-Pass Editing}
\label{sec: night_details}

\subsection{BRDF Modeling}
Our BRDF model combines diffuse and specular reflections under multiple spotlight sources with inverse-square falloff:

\begin{equation}
L_{\text{out}}(\mathbf{x}, \mathbf{\omega}_o) = \int_{\Omega} f_r(\mathbf{x}, \mathbf{\omega}_i, \mathbf{\omega}_o) L_i(\mathbf{x}, \mathbf{\omega}_i) (\mathbf{n} \cdot \mathbf{\omega}_i) \, d\mathbf{\omega}_i
\end{equation}

where $\mathbf{x}\in\mathbb{R}^3$ is the world-space position of the surface point being shaded, $f_r$ is the Cook-Torrance BRDF combining metallic-roughness material properties, $\mathbf{\omega}_o$ and $\mathbf{\omega}_i \in \mathbb{S}^2$ are unit vectors representing view and incident light directions respectively ($\mathbf{\omega}_i$ is direction of incoming radiance, pointing toward the surface; $\mathbf{\omega}_o$ is direction of reflected radiance), $\mathbf{n}$ is the surface normal, and $\Omega$ is the upper hemisphere oriented by $\mathbf{n}$.

The Cook-Torrance BRDF combines diffuse and specular terms:
\begin{equation}
f_r = \frac{\mathbf{c}_{\text{diff}}}{\pi}(1 - m) + \frac{D\cdot G\cdot F}{4(\mathbf{n} \cdot \mathbf{\omega}_o)(\mathbf{n} \cdot \mathbf{\omega}_i)}
\end{equation}
where $\mathbf{c}_{\text{diff}}$ is the diffuse color coefficient (surface albedo), $m$ is the metallic parameter, and the specular term consists of the following components computed using the Filament/Disney PBR convention. As these components follow well-established definitions in the field, we omit redundant details of notation and present the core formulations below:
\begin{itemize}
    \item Roughness remapping: $\alpha = r^2$ where $r$ is perceptual roughness
    \item GGX distribution~\cite{10.5555/2383847.2383874}: $D = \frac{\alpha^2}{\pi((\mathbf{n} \cdot \mathbf{h})^2(\alpha^2 - 1) + 1)^2}$
    \item Smith height-correlated visibility~\cite{Smith1967GeometricalSO}:
    \[G = \frac{1}{2(\lambda_o + \lambda_i)}\]
    where $\lambda_o$ and $\lambda_i$ represent the microfacet shadowing and masking terms for the outgoing and incoming directions,
    \begin{align*}
    \lambda_o &= (\mathbf{n} \cdot \mathbf{\omega}_i)\sqrt{(\mathbf{n} \cdot \mathbf{\omega}_o)^2(1-\alpha^2) + \alpha^2}\\
    \lambda_i &= (\mathbf{n} \cdot \mathbf{\omega}_o)\sqrt{(\mathbf{n} \cdot \mathbf{\omega}_i)^2(1-\alpha^2) + \alpha^2}
    \end{align*}
    \item Schlick Fresnel~\cite{Schlick1994AnIB}: $F = F_0 + (1 - F_0)(1 - \mathbf{\omega}_i \cdot \mathbf{h})^5$
\end{itemize}
where $\mathbf{h} = \frac{\mathbf{\omega}_i + \mathbf{\omega}_o}{||\mathbf{\omega}_i + \mathbf{\omega}_o||}$ is the halfway vector and $F_0 = \text{lerp}(0.04, \text{albedo}, \text{metallic})$ is the Fresnel reflectance at normal incidence.

\subsection{Light Source Implementation}
\textbf{Incident Radiance \& Spotlight Modeling.} The incident radiance $L_i$ aggregates attenuated contributions from all discrete light sources:

\begin{equation}
L_i(\mathbf{x}, \mathbf{\omega}_i) \approx \sum_{j} \frac{\mathbf{E}_j \cdot A_j(\mathbf{x})}{||\mathbf{x} - \mathbf{p}_j||^2 + \epsilon}
\end{equation}

where $\mathbf{E}_j \in \mathbb{R}^3$ is the RGB radiant intensity (in linear space) of light source $j$, $\mathbf{p}_j$ is its 3D position, $\epsilon$ prevents division by zero, and $A_j(\mathbf{x})$ represents combined angular and distance attenuation factors. 

Street lights and vehicle headlights are modeled as spotlights with both angular $S_j$ and $W_j$ distance attenuation. The combined attenuation factor is:
\begin{equation}
A_j(\mathbf{x}) = S_j(\mathbf{x}) \cdot W_j(\mathbf{x})
\end{equation}

For angular attenuation, we implement a smooth falloff between inner and outer cone angles:
\begin{equation}
S_j(\mathbf{x}) = \left(\frac{\max(0, \cos\theta_j - \cos\theta_{\text{outer}})}{\cos\theta_{\text{inner}} - \cos\theta_{\text{outer}}}\right)^2
\label{eq:light_pass}
\end{equation}

where $\theta_j$ is the angle between the spotlight's forward direction $\mathbf{d}_j$ and the vector from light to surface $(\mathbf{x} - \mathbf{p}_j)$. Typical values: street lights use $\theta_{\text{inner}}=15^{\circ}$, $\theta_{\text{outer}}=35^{\circ}$; vehicle headlights use $\theta_{\text{inner}}=10^{\circ}$, $\theta_{\text{outer}}=25^{\circ}$.

To ensure finite computational domains and physically plausible falloff:
\begin{equation}
W_j(\mathbf{x}) = \left(1 - \left(\frac{||\mathbf{x} - \mathbf{p}_j||}{r_{\text{max}}}\right)^4\right)^2_{+}
\end{equation}
where $r_{\text{max}}$ is the light's influence radius (typically 10-20 meters for street lights, 30-50 meters for headlights), and $(\cdot)_+$ denotes clamping negative values to zero.

\subsection{LUT Construction for Global Tone Mapping}
\label{sec:lut_construction}

As mentioned in the main paper, we apply global tone mapping via a parametric Look-Up Table (LUT) to achieve realistic nocturnal appearance. The LUT simultaneously darkens ambient regions and shifts color temperature to warm tones characteristic of night scenes, while preserving visibility in areas without artificial lighting.

\noindent\textbf{LUT Parameterization.}
We construct a LUT as a $256 \times 3$ array mapping input RGB intensities to output values. For nocturnal tone mapping with strength parameter $\sigma \in [0, 1]$, each input intensity $i \in [0, 255]$ is transformed as:
\begin{equation}
\begin{aligned}
\beta &= 0.7 + 0.2(1 - \sigma) \\
R_{\text{out}} &= \text{clip}(\beta \cdot i \times (0.85 - 0.15\sigma), 0, 255) \\
G_{\text{out}} &= \text{clip}(\beta \cdot i \times (0.9 - 0.1\sigma), 0, 255) \\
B_{\text{out}} &= \text{clip}(\beta \cdot i \times (1.05 + 0.2\sigma), 0, 255)
\end{aligned}
\end{equation}
where $\beta$ controls overall brightness reduction while preserving detail in dark regions. The channel-specific scaling creates the cool blue tones of moonlight while maintaining warm artificial light sources.

\noindent\textbf{Adaptive Exposure Pre-processing.}
Before applying the LUT, we perform per-frame adaptive exposure adjustment in linear color space to prevent over-darkening. After converting from sRGB to linear space, we estimate scene luminance $L_p$ at the 70-th percentile and compute an adaptive gain:
\begin{equation}
\gamma = (0.98 - 0.20\sigma) \times \text{clip}\left(\frac{0.22}{L_p + \epsilon}, 0.6, 1.6\right)
\end{equation}
We apply this gain in linear space, followed by highlight compression $C' = C/(1 + 0.25C)$ before converting back to sRGB for LUT application.

\noindent\textbf{Sky Region Handling.}
To prevent unrealistic brightening of sky regions, we apply differential processing using segmented sky masks. Sky pixels are darkened by a factor $\alpha_{\text{sky}} = 0.6$, with soft boundaries created by dilating the mask by 20 pixels and applying Gaussian blur ($\kappa = 10$). This ensures natural transitions at horizon boundaries while maintaining the dark night sky appearance.

\section{Volumetric Fog Synthesis via G-Buffer Dual-Pass Editing}
\label{sec:fog_details}
\subsection{Fog modeling}
We model atmospheric scattering via a simplified radiative transfer equation (RTE) under uniform medium assumptions. For a view ray $\mathbf{r}(t) = \mathbf{o} + t\mathbf{d}$ reaching a surface at depth $s$, the observed radiance is:
\begin{equation}
L_{\text{obs}} = L_{\text{surface}} \cdot T(s) + L_{\text{in-scatter}},
\end{equation}
where $L_{\text{surface}}$ is the surface radiance computed via the Cook-Torrance BRDF model, $T(s) = \exp(-\sigma_t \cdot s)$ is the transmittance with extinction coefficient $\sigma_t = \sigma_a + \sigma_s$ (absorption $\sigma_a$ and scattering $\sigma_s$), and $L_{\text{in-scatter}}$ is the accumulated in-scattered light.

The in-scattering term sums contributions from all light sources $\mathcal{L}$:
\begin{equation}
L_{\text{in-scatter}} = \sum_{i \in \mathcal{L}} \sigma_s \cdot p(\mathbf{d}, \mathbf{d}_i) \cdot L_i \cdot \gamma,
\end{equation}
where $p(\mathbf{d}, \mathbf{d}_i)$ is the Henyey-Greenstein phase function ~\cite{Henyey1940DiffuseRI} with forward-scattering parameter $g=0.8$, $L_i$ is the attenuated light intensity, and $\gamma$ is the scattering strength. Both $\sigma_s$ and $\sigma_a$ are scaled by a density factor $\alpha$ for real-time control.

For more specifically, the Henyey-Greenstein phase function used in our fog model is:
\begin{equation}
p(\mathbf{d}, \mathbf{d}_i) = \frac{1 - g^2}{4\pi (1 + g^2 - 2g \cos\theta)^{3/2}},
\end{equation}
where $\theta$ is the scattering angle between view direction $\mathbf{d}$ and light direction $\mathbf{d}_i$. We use $g=0.8$ to model forward scattering typical of fog particles.

For different light types (directional, point, spot), we compute appropriate attenuation factors, including distance falloff and angular attenuation for spot lights. The density scaling parameters $\sigma_s' = \sigma_s \cdot \alpha$ and $\sigma_a' = \sigma_a \cdot \alpha$ allow real-time adjustment of fog density without recomputation.

\subsection{Fog blending}
The final color blending for artistic control is computed as:
\begin{equation}
L_{\text{final}} = (1 - \beta \cdot f) \cdot L_{\text{obs}} + \beta \cdot f \cdot \mathbf{F},
\end{equation}
where $\mathbf{F}$ is the fog color, $f = 1 - T(s)$ represents the fog opacity, and $\beta$ controls the blend strength, we set it as 0.5.

\section{Environment Harmonization}
\label{sec: environment_light_pass_editing}

To achieve natural integration of ambient light and directional/local illumination, we adopt an adaptive linear blending strategy operating in the linear light space (rather than sRGB) to avoid gamma correction-induced distortion of light intensity relationships. The fusion pipeline begins with converting both ambient light maps and directional/local illumination maps from sRGB to linear space, as sRGB’s non-linear encoding would misrepresent the additive physical interactions between ambient and direct light components.  

For adaptive weighting, we first compute the per-pixel illuminance mean of the directional/local illumination map (across RGB channels), which quantifies the intensity of direct light contributions at each spatial position. To suppress trivial or noisy direct light signals (e.g., low-intensity artifacts), we clamp this illuminance value to a valid range ([0, 1]) with a lower threshold (0.05), ensuring only meaningful direct illumination regions influence the blend. This clamped illuminance is further scaled by a tunable strength factor to control the overall weight of directional/local lighting, yielding an adaptive weight map $W_{\text{direct}}$.  

The final fusion is performed via linear combination in linear space: \begin{equation}
I_{\text{blended}} = (1 - W_{\text{direct}}) \cdot I_{\text{ambient, linear}} + W_{\text{direct}} \cdot I_{\text{direct, linear}}
\end{equation} 
Here, $ I_{\text{ambient, linear}} $ denotes the ambient light map (capturing global illumination, which is derived from the output of the pretrained neural forward renderer~\cite{DiffusionRenderer}) and $ I_{\text{direct, linear}} $ represents directional/local illumination (e.g., point light sources, spotlights) in linear space. This formulation ensures that regions with stronger direct illumination (e.g., streetlight highlights) are dominated by $I_{\text{direct}}$, while areas lacking direct light retain the ambient light background—preserving the natural contrast between global ambient filling and localized direct lighting. After blending, the result is converted back to sRGB space for display consistency.  

This approach adheres to the physical plausibility of light mixing (critical for photorealism) and enables spatially adaptive fusion that respects the intensity hierarchy between ambient and direct illumination, avoiding over-saturation or unnatural transitions in the final rendered scene.

\section{Text Prompt Design and Templates}
\label{sec:prompt}

The final prompt is constructed as a combination of a base prompt and a conversion instruction. 

The base prompt is generated using Qwen3-8B-Instruct~\cite{bai2025qwen3}, by feeding the original video frame and asking the model to describe the video. The description is required to cover: (1) road layout and surrounding elements such as trees and barriers; (2) objects including vehicles, utility poles, wires, and billboards; (3) motion state and traffic density; (4) overall scene atmosphere; (5) the state of traffic lights and traffic signs, if visible. We constrain the description to 2–4 sentences and follow this example style:

``Several cars are visible on the road, some moving forward while others appear stationary, indicating moderate traffic. The road is flanked by trees and a concrete barrier on one side, with utility poles and wires running parallel to the highway. A billboard is visible in the distance, and the overall atmosphere suggests a calm urban or suburban setting.''

For the conversion instruction, we design dedicated text prompts for each target scenario to guide the diffusion model toward generating physically plausible and visually appealing weather and time-of-day effects, with an emphasis on cinematic quality, fine-grained physical details, and semantic consistency.

\subsection{Night Scene Conversion Instruction}\textit{Instruction:} A highly realistic, cinematic midnight street scene. The sky is pitch black, set in the deep hours of midnight. Vehicles have their headlights turned on, illuminating the dark road. Streetlights cast warm glows, but most of the scene is enveloped in darkness. Photographic clarity, true-to-life colors, realistic shadows, no artificial effects, just a believable urban night.

\textit{Design Rationale:} Focuses on time specificity (midnight) and light-source interactions (headlights, streetlights). ``Pitch black sky'' and ``enveloped in darkness'' constrain overall brightness, while ``photographic clarity'' and ``realistic shadows'' ensure physical consistency for an authentic urban night atmosphere.
\subsection{Fog Scene Conversion Instruction}\textit{Instruction:} A realistic daytime street scene in thick fog. Vehicles are driving through dense mist with their headlights turned on, illuminating the foggy road. Diffused daylight filtering through the fog, low visibility and strong atmospheric perspective, subtle vehicle headlights and streetlight glows visible through haze, realistic volumetric lighting and photographic detail, believable urban environment.

\textit{Design Rationale:} Highlights atmospheric optical effects (volumetric light, atmospheric perspective in fog) and light scattering (faint glows of headlights/streetlights in haze). ``Low visibility'' and ``subtle glows through haze'' ensure fog opacity and light-scattering behavior align with real-world physics.
\subsection{Snowy Scene Conversion Instruction}\textit{Instruction:} A highly realistic, cinematic snowy street scene. trees, buildings, and vehicles covered in snow, irregular patches of accumulated snow on the road surface, gently falling snowflakes in the air, realistic shadows and reflections on wet surfaces, photographic clarity, true-to-life colors, no artificial effects, just a believable urban winter environment.

\textit{Design Rationale:} Addresses snow accumulation details (irregular road snow patches, snow coverage on objects) and dynamic elements (falling snowflakes). ``Reflections on wet surfaces'' reflect the physical properties of partially melted snow, while ``cinematic'' and ``photographic clarity'' balance artistic presentation and visual realism.
\subsection{Rainy Scene Conversion Instruction}\textit{Instruction:} A highly realistic, cinematic rainy street scene. Overcast sky with diffused light, rain-soaked roads and sidewalks. The rainwater on the ground formed ripples, realistic raindrops, photographic clarity, true-to-life colors, no artificial effects, just a believable urban rainy environment.

\textit{Design Rationale:} Emphasizes hydrological details (wet roads, rain ripples) and lighting conditions (overcast diffuse light). ``Realistic raindrops'' and ``no artificial effects'' ensure rain morphology and motion follow natural laws, with ``cinematic'' style balancing artistic texture and physical accuracy.

\section{VidRefiner Architecture and Configurations}
\label{sec:postprocess}
In the postprocessing stage, we fine-tune several parameters to balance preservation of physical edits and artifact repair. The editing strength $\alpha$ is typically set to 0.4, injecting moderate noise to refine details without overwriting the underlying weather simulation. Classifier-free guidance scale is set to $\gamma = 6$ to emphasize prompt adherence. The diffusion process uses $T = 20$ inference steps, with default scheduler in WAN-FUN 2.2-5B~\cite{wan2025}. Also for efficiency, we clip the video to resolution 1280$\times$704.

Pseudocode for the collapsed searching space generation scheme -proven in~\cite{meng2022sdedit}- is as follows, where $v$ denotes the input video, $\alpha \in [0, 1]$ denotes the editing strength controlling the magnitude of injected noise (a higher $\alpha$ injects more noise and allows for stronger topological deviations), $T$ the total number of diffusion timesteps, $z$ the latent representation, $t_s$ the starting pivot timestep for denoising, $\sigma_t$ the noise schedule at timestep $t$, $\epsilon$ the Gaussian noise, $\hat{\epsilon}$ the predicted noise, $cond$ for the condition video, and $\hat{v}$ the refined output video:

\begin{algorithm}
\caption{Collapsed searching space generation scheme}
\SetKwFunction{FPostprocess}{postprocess}
\SetKwProg{Fn}{Function}{:}{}
\Fn{\FPostprocess{v, prompt, $\alpha$, $T$}}{
    $z \gets \text{encode}(v)$ \tcp{Encode to latent}
    $t_s \gets \lfloor (T - 1) \times \alpha \rfloor$ \tcp{Start step}
    $z_{t_s} \leftarrow \sqrt{\bar{\alpha}_{t_s}}z + \sqrt{1-\bar{\alpha}_{t_s}}\epsilon, \quad \epsilon \sim \mathcal{N}(0, I) \quad // \text{Add noise up to timestep } t_s$
    \For{$t = t_s$ \KwTo $0$}{
        $\hat{\epsilon} \gets \text{predict}(z_t, prompt, t, cond)$ \tcp{Predict noise}
        $z_t \gets \text{denoise}(z_t, \hat{\epsilon}, t, cond)$ \tcp{One step denoise}
    }
    $\hat{v} \gets \text{decode}(z_t)$ \tcp{Decode to video}
    \Return{$\hat{v}$}\;
}
\end{algorithm}

\section{Extended Quantitative Evaluations}
\label{sec:more_quantitative}

\textbf{AABB Bounding-Box Projection and IoU Formulation}

To strictly quantify the structural consistency mentioned in the main text, we evaluate the 2D Intersection-over-Union (IoU) between the Axis-Aligned Bounding Boxes (AABB). To ensure a strictly fair comparison, both the Ground Truth (GT) boxes and the predicted boxes extracted by the OvMono3D~\cite{OVMono3D} detector are projected from 3D space onto the 2D image plane using the exact same mathematical formulation.

For any given 3D bounding box (either the dataset GT or the OvMono3D prediction), let its 8 corners in the camera coordinate system be denoted as $P_i = [X_i, Y_i, Z_i]^T$, where $i \in \{1, \dots, 8\}$. We project these 3D corners onto the 2D image plane using the camera intrinsic matrix $K$:

\begin{equation}
\begin{bmatrix} x_i \\ y_i \\ z_i \end{bmatrix} = K \begin{bmatrix} X_i \\ Y_i \\ Z_i \end{bmatrix}
\end{equation}

The 2D pixel coordinates are then obtained by perspective division:

\begin{equation}
u_i = \frac{x_i}{z_i}, \quad v_i = \frac{y_i}{z_i}
\end{equation}

To form the 2D Axis-Aligned Bounding Box (AABB) $B$, we compute the minimum enclosing rectangle of these 8 projected points:

\begin{equation}
B=\{ (u,v) \mid u_{min} \le u \le u_{max}, v_{min} \le v \le v_{max} \}
\end{equation}

where $u_{min}=\min_{i}(u_i)$, $u_{max}=\max_{i}(u_i)$, $v_{min}=\min_{i}(v_i)$, and $v_{max}=\max_{i}(v_i)$.

Finally, given the projected GT box $B_{gt}$ and the projected predicted box $B_{pred}$, the structural consistency is measured via the standard 2D IoU:

\begin{equation}
    IoU=\frac{Area(B_{gt} \cap B_{pred})}{Area(B_{gt} \cup B_{pred})}
\end{equation}

\textit{Note: This metric serves as a strict proxy for structural preservation. A drop in IoU may encompass both actual geometric degradation during the generative process and the natural performance degradation of the pre-trained 3D detector (OvMono3D) when confronted with severe, out-of-distribution adverse weather conditions.}

As analyzed in the main paper, our method achieves the best CLIP score, and best vehicle geometric and perceptual alignment, demonstrating the strongest instruction adherence among all baseline approaches.

To quantify the overall consistency between original and edited videos, we evaluate depth and edge alignment across 120 sequences covering four common weather and time-of-day scenarios: fog, rain, snow, and night.

\textbf{Depth Alignment} (Depth si-RMSE):
We employ DepthAnythingV2~\cite{video_depth_anything} to extract depth maps from both input videos and generated results, and compute the scale-invariant root mean square error (si-RMSE)~\cite{10.5555/2969033.2969091} for evaluation. Lower values denote better depth alignment.

\begin{table}[ht]
\centering
\caption{\textbf{Depth alignment evaluation.} Depth si-RMSE ($\downarrow$) averaged over 480 generated videos.}
\label{tab:depth_alignment}
\begin{tabular}{l|c}
\toprule
Model & Depth si-RMSE $\downarrow$ \\
\midrule
Video-P2P~\cite{liu2023videop2p} & 0.511 \\
Ditto~\cite{bai2025ditto} & 0.632 \\
Cosmos-Transfer2.5~\cite{nvidia2025worldsimulationvideofoundation} & 0.225 \\
WAN-FUN 2.2~\cite{wan2025} & 0.267 \\
Ours & 0.247 \\
\bottomrule
\end{tabular}
\end{table}

The minor drop in depth alignment mainly stems from the explicit depth we generate for falling snow and rain, whereas most baselines cannot produce geometrically consistent snow or rain at all.

\textbf{Edge Alignment} (Edge F1): We apply Canny edge extraction~\cite{4767851} (with low threshold $t_1=30$ and high threshold $t_2=60$) to both videos and calculate the F1 score for black-and-white pixel classification. Higher values indicate better alignment.

\begin{table}[ht]
\centering
\caption{\textbf{Edge alignment evaluation.} Edge F1 scores (↑) for different methods averaged over 480 generated videos.}
\label{tab:edge_f1}
\begin{tabular}{l|c}
\toprule
Model & Edge F1 $\uparrow$ \\
\midrule
Video-P2P \cite{liu2023videop2p} & 0.104 \\
Ditto \cite{bai2025ditto} & 0.043 \\
Cosmos-Transfer2.5 \cite{nvidia2025worldsimulationvideofoundation} & 0.321 \\
WAN-FUN 2.2 \cite{wan2025} & 0.187 \\
Ours & 0.129 \\
\bottomrule
\end{tabular}
\end{table}

To clarify the quantitative behavior on the Edge F1 metric, it is important to note that our method performs explicit geometry editing for weather synthesis (e.g., snow accumulation altering surface, puddles perturbing normals, wetness changing reflectance, and fog/night illumination reshaping edge contrast). These geometry‑aware edits intentionally modify scene structure, producing rendered frames whose silhouettes and fine‑scale edges differ from the original input video. In contrast, Cosmos‑Transfer2.5 and WAN‑FUN2.2 obtain edges from the original video and use them as control input, meaning their edited outputs remain closer to the original edge distribution. Because the Edge F1 metric uses Canny edges extracted from the original video as ground truth, methods that preserve input geometry naturally achieve higher scores. Our pipeline, however, generates physically based modified geometry and lighting, so the ``ground truth'' edges—taken from the unedited video—no longer match the altered structures. This mismatch leads to a lower Edge F1 despite higher physical realism. Therefore, the metric reflects structural deviation rather than degradation, and the drop in Edge F1 is an expected outcome of our physically grounded geometry modification rather than a failure of edge preservation.

We further report two perceptual metrics evaluated on the generated videos.

\textbf{Temporal Consistency} (Fréchet Video Distance)~\cite{Unterthiner2019FVDAN}:
This metric measures the distributional similarity between real and generated videos, characterizing temporal coherence and visual realism.
We evaluate on four weather-specific datasets, each containing 120 videos (57 frames per video).
The FVD score is computed separately for each dataset and averaged to produce the final result.

\textbf{Overall Quality}:
We adopt the DOVER (Disentangled Objective Video Quality Evaluator) score~\cite{wu2023dover} as our perceptual quality metric.
DOVER is a learning-based video quality assessment framework that disentangles quality evaluation into two complementary components:
\begin{itemize}
    \item \textit{Aesthetic perspective} ($Q_{\text{pred,A}}$): Processes spatially resized frames (224×224) to preserve semantic content while reducing sensitivity to low-level distortions.
    \item \textit{Technical perspective} ($Q_{\text{pred,T}}$): Analyzes local spatiotemporal patches to detect technical artifacts while being invariant to global aesthetic composition.
\end{itemize}
The final score is a weighted combination of the two:
\begin{equation}
Q_{\text{pred}} = 0.428Q_{\text{pred,A}} + 0.572Q_{\text{pred,T}},
\end{equation}
where we use DOVER++ weights for scoring and report the average over the dataset.

\begin{table}[ht]
\centering
\caption{\textbf{Perceptual metric evaluation.} Temporal consistency (FVD) and visual quality (DOVER) averaged over 480 generated videos. \textit{Higher DOVER scores} indicate better visual quality, while \textit{lower FVD scores} indicate better temporal consistency.}
\label{tab:perceptual_metrics}
\begin{tabular}{l|cc}
\toprule
Model & FVD Score $\downarrow$ & DOVER Score $\uparrow$ \\
\midrule
Video-P2P~\cite{liu2023videop2p} & 1687.4 & 0.194 \\
Ditto~\cite{bai2025ditto} & 2024.2 & 0.345 \\
Cosmos-Transfer2.5~\cite{nvidia2025worldsimulationvideofoundation} & 808.7 & 0.358 \\
WAN-FUN 2.2~\cite{wan2025} & 1029.5 & 0.448 \\
Ours & 886.8 & 0.370 \\
\bottomrule
\end{tabular}
\end{table}

Notably, these two perceptual metrics do not fully reflect editing \textit{correctness}, as some baselines can generate visually appealing results yet fail to strictly adhere to the given editing instructions. Thus, they serve only as a reference.

\section{Comprehensive ablation studies}
\label{sec: ablation_studies}

In this section, we present a comprehensive set of ablation studies to validate our framework. This includes: (1) single-module effectiveness analysis, (2) pair-wise and multi-module synergy investigations, and (3) intra-module integration evaluations.

Evaluation Protocol: Module Sensitivity Analysis via Spatial Divergence. It is important to explicitly clarify our evaluation methodology for these internal ablations. In the absence of paired real-world ground truth for counterfactual weather conditions, we utilize the deterministic output of our full pipeline (with all components active) as a reference anchor. By computing the PSNR between the ablated variants and this full-pipeline anchor, we conduct a Module Sensitivity Analysis.

We emphasize that in this specific context, this internal PSNR is not employed as a metric of perceptual quality or physical realism. Instead, we repurpose PSNR strictly as a pixel-level spatial divergence metric to quantitatively isolate the magnitude of intervention for each specific module. A lower internal PSNR indicates a higher structural and photometric deviation from the final designed output, explicitly demonstrating that the ablated module contributes substantially to the physical simulation process. This metric is designed solely to quantify the deterministic ``degree of alteration'' introduced by individual modules, validating their functional necessity within our dual-pass editing mechanism.

\begin{table*}[htbp]
  \centering
  \caption{Ablation Study on Single Module Effectiveness}
  \label{tab:ablation_single_module}
  \begin{tabular}{@{}l p{5cm} p{6cm}@{}}
    \toprule
    \textbf{ID} & \textbf{Core Verification Point} & \textbf{Experimental Configuration Design} \\
    \midrule
    \ding{172} 
    & 4D reconstruction can effectively prevent aliasing in local light rendering 
    & Experiment A: Compare the effect with and without the 4D reconstruction module in Fig.~\ref{fig:ab_A}. \\
    \ding{173} 
    & Inverse rendering + Env light Editing can effectively manipulate shadows according to HDR changes 
    & Experiment B: Compare shadow manipulation results with and without the combination of Inverse rendering + Env light Editing in Fig.~\ref{fig:ab_B}. \\
    \ding{174} 
    & Geometry pass editing can effectively handle snow/rain weather editing based on prompts 
    & Experiment C: Compare rain editing results with/without Geometry pass editing, and snow editing results with/without Geometry pass editing in Fig.~\ref{fig:ab_C}. \\
    \ding{175} 
    & Light pass editing can effectively handle fog/night weather editing 
    & Experiment D: Compare fog editing results with/without Light pass editing, and night editing results with/without Light pass editing in Fig.~\ref{fig:ab_D}. \\
    \ding{176} 
    & VidRefiner can effectively remove noise in G-Buffer 
    & Experiment E: Compare noisy G-Buffer processing results with and without VidRefiner in Fig.~\ref{fig:ab_E}. \\
    \bottomrule
  \end{tabular}
\end{table*}

\subsection{Single-module ablation studies}
To systematically validate the independent contributions of each core module in our framework, we first conduct a series of single-module ablation studies. These experiments aim to isolate the functionality of each component, quantifying its specific role in enhancing the overall performance without confounding effects from other modules. Each module is designed to address a distinct task within the pipeline, and our ablation strategy focuses on comparing results with and without the target module to explicitly measure its effectiveness. The core verification point and experimental configuration design is shown in Tab.~\ref{tab:ablation_single_module}.

\begin{figure*}[t]
  \centering
    \includegraphics[width=1.0\linewidth]{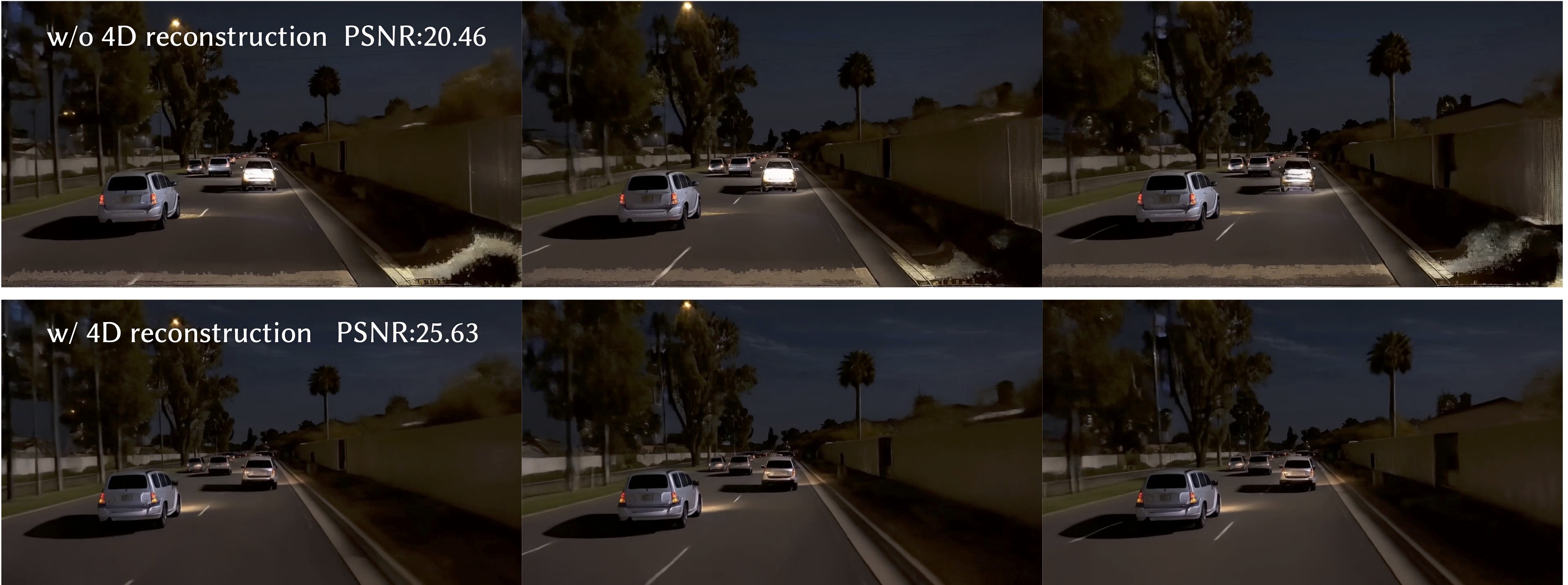}
    \caption{Experiment A: Compare the effect with and without the 4D reconstruction module.
}
    \label{fig:ab_A}
\end{figure*}

\begin{figure*}[t]
  \centering
    \includegraphics[width=1.0\linewidth]{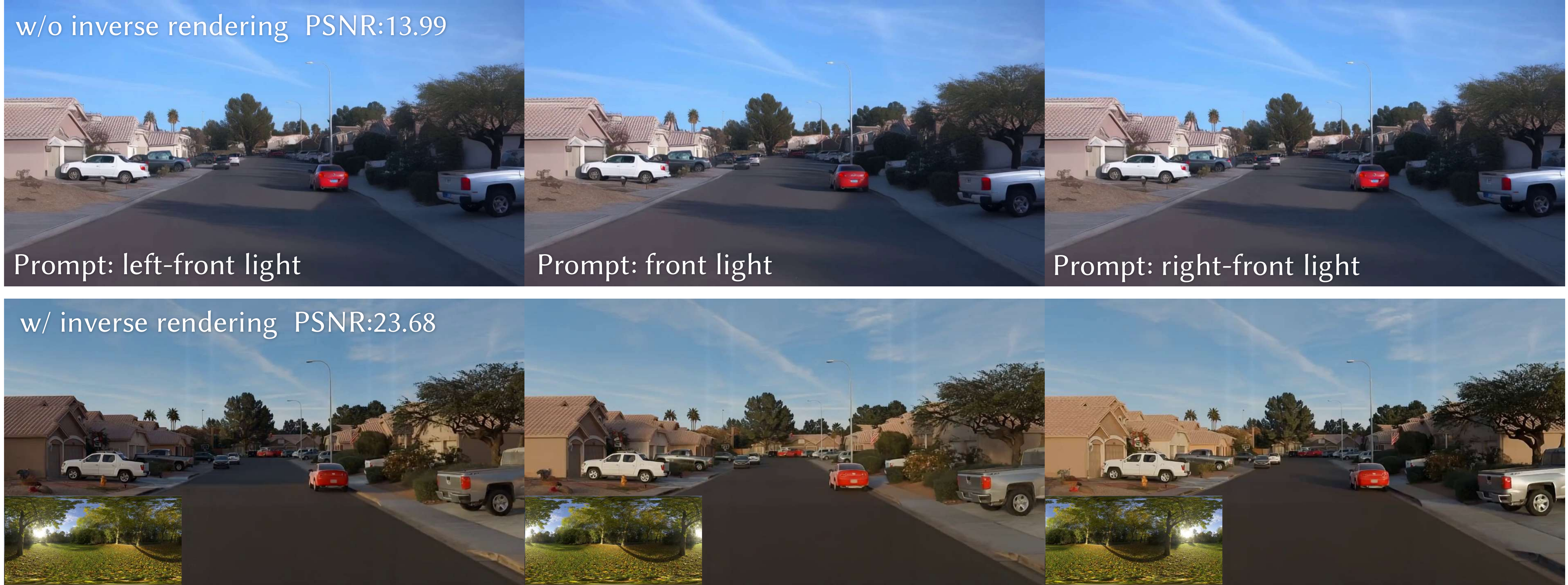}
    \caption{Experiment B: Compare shadow manipulation results with and without the combination of Inverse rendering + Env light Editing.
}
    \label{fig:ab_B}
\end{figure*}

\begin{figure*}[t]
  \centering
    \includegraphics[width=1.0\linewidth]{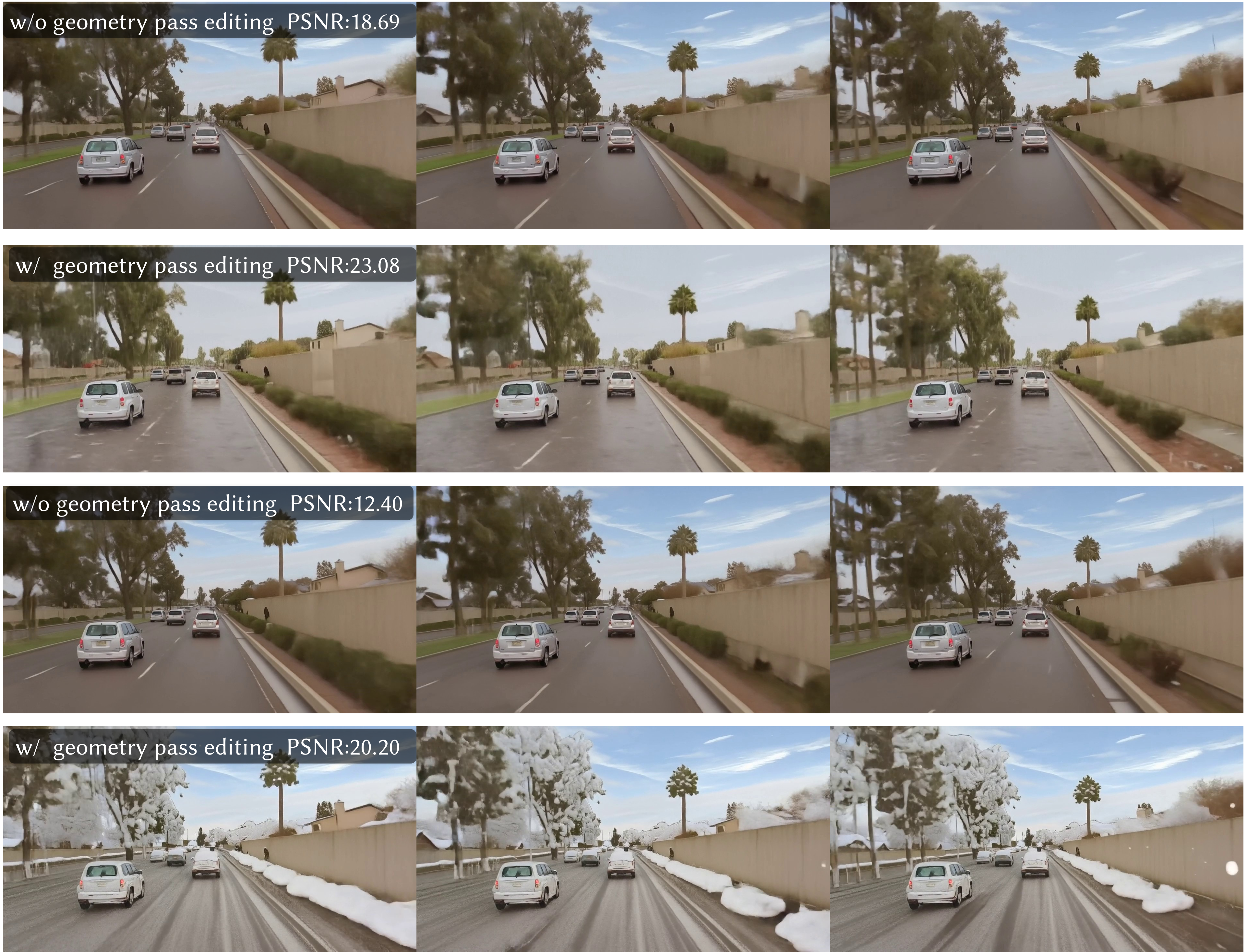}
    \caption{Experiment C: Compare rain editing results with/without Geometry pass editing, and snow editing results with/without Geometry pass editing.
}
    \label{fig:ab_C}
\end{figure*}

\begin{figure*}[t]
  \centering
    \includegraphics[width=1.0\linewidth]{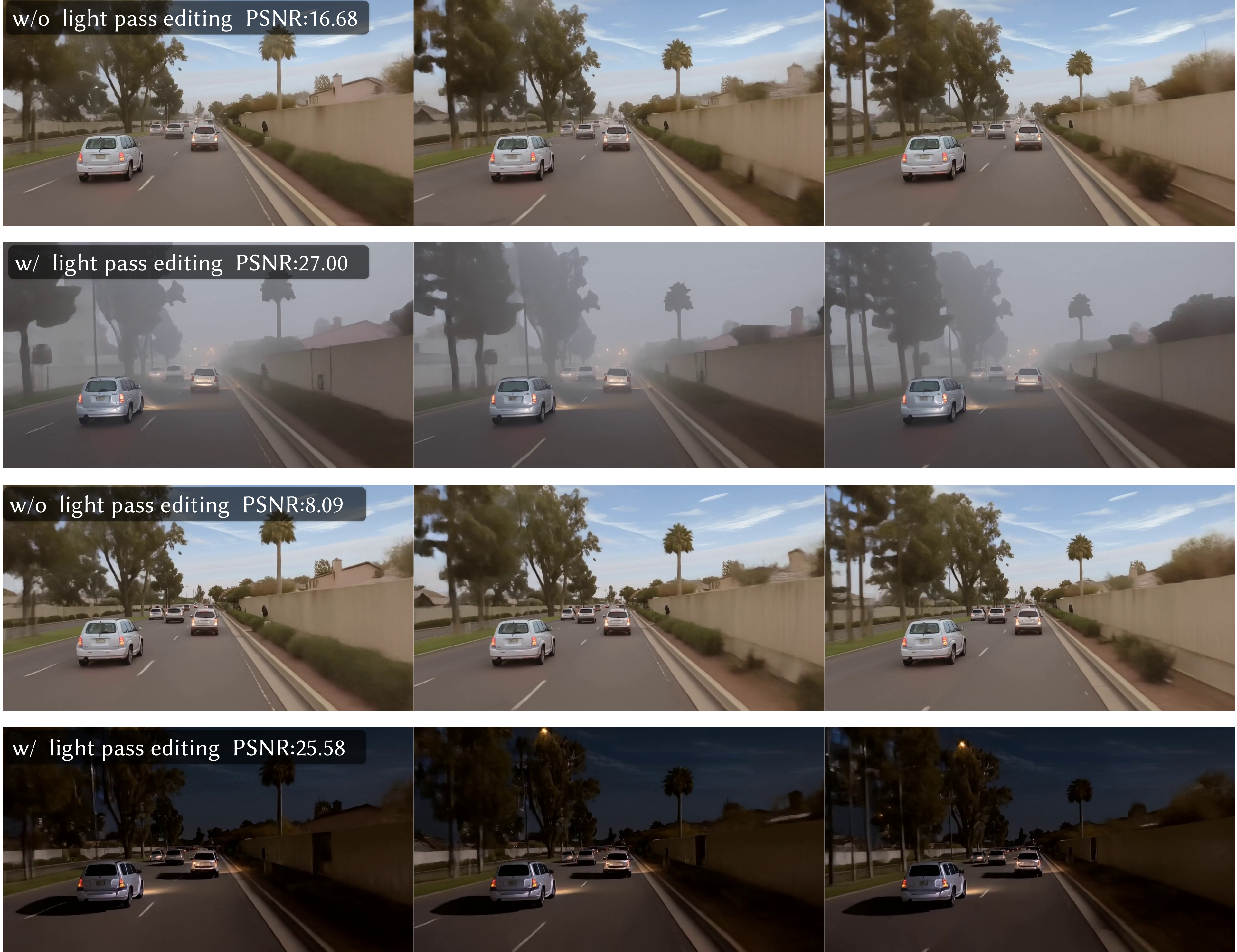}
    \caption{Experiment D: Compare fog editing results with/without Light pass editing, and night editing results with/without Light pass editing.
}
    \label{fig:ab_D}
\end{figure*}

\begin{figure*}[t]
  \centering
    \includegraphics[width=1.0\linewidth]{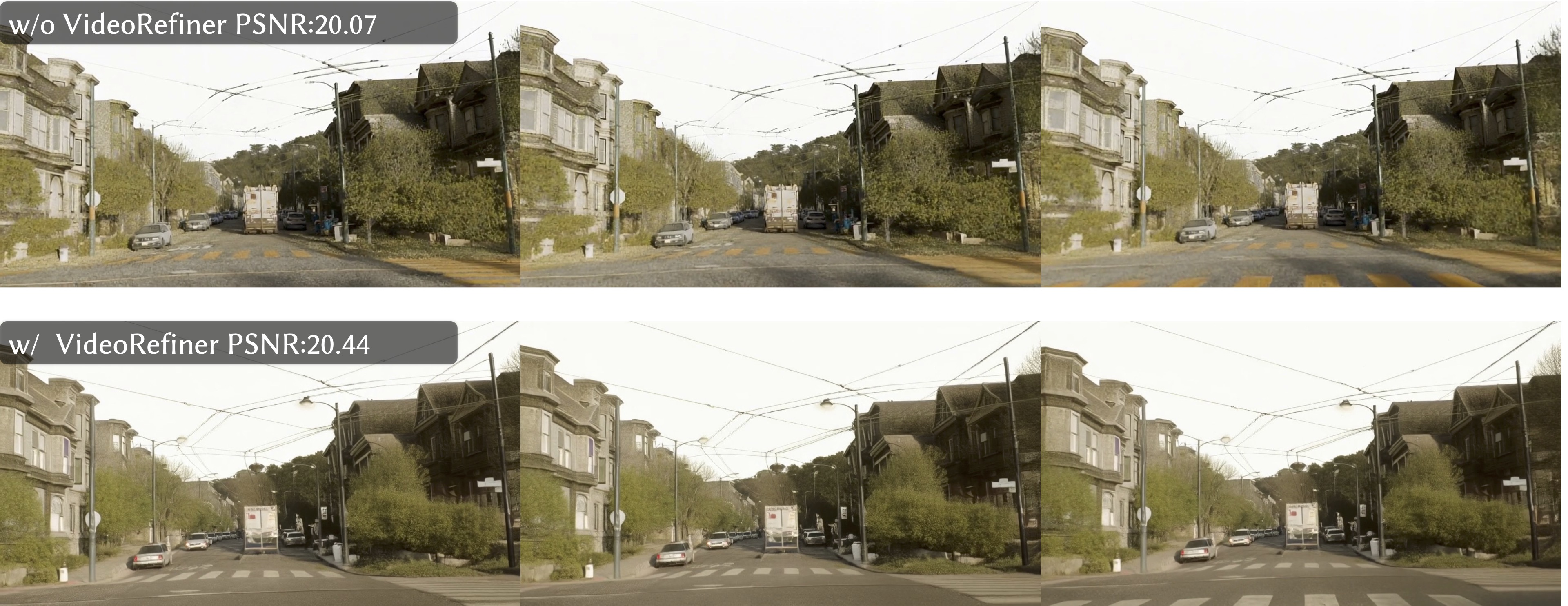}
    \caption{Experiment E: Compare noisy G-Buffer processing results with and without VidRefiner.
}
    \label{fig:ab_E}
\end{figure*}

\clearpage

\subsection{Synergistic-modules ablation studies}

\begin{table*}[htbp]
  \centering
  \caption{Ablation Study on Module Synergies (VidRefiner + Functional Modules)}
  \label{tab:syn_module_1}
  \begin{tabular}{@{}l p{5.5cm} p{5.5cm}@{}}
    \toprule
    \textbf{ID Pair} & \textbf{Core Verification Point} & \textbf{Experimental Configuration Design} \\
    \midrule
    \ding{172} + \ding{176} 
    & Whether VidRefiner enhances fidelity of anti-aliased local light rendering 
    & Experiment F: Compare local light rendering results of ``4D Recon alone'' vs. ``4D Recon + VidRefiner'' in Fig.~\ref{fig:ab_F}. \\
    \ding{173} + \ding{176} 
    & Whether VidRefiner preserves shadow manipulation accuracy while enhancing fidelity 
    & Experiment G: Compare shadow editing results of ``Inverse Renderer+Env Light Editing alone'' vs. ``the combination + VidRefiner'' in Fig.~\ref{fig:ab_G}. \\
    \ding{174} + \ding{176} 
    & Whether VidRefiner maintains weather effect realism and improves frame fidelity 
    & Experiment H: Compare snow/rain editing results of ``Geometry Pass Editing alone'' vs. ``Geometry Pass Editing + VidRefiner'' in Fig.~\ref{fig:ab_H}. \\
    \ding{175} + \ding{176} 
    & Whether VidRefiner enhances fidelity of fog/night lighting edits 
    & Experiment I: Compare fog/night editing results of ``Light Pass Editing alone'' vs. ``Light Pass Editing + VidRefiner'' in Fig.~\ref{fig:ab_I}. \\
    \bottomrule
  \end{tabular}
\end{table*}

Because \ding{172}–\ding{175} in Tab.~\ref{tab:ablation_single_module} are independent modules with non-overlapping core functionalities, \ding{176} (VidRefiner) serves as a general-purpose, universally compatible optimization module that can be arbitrarily combined with any subset of \ding{172}–\ding{175} to enhance video spatial fidelity. To verify the synergistic gains between VidRefiner and these functional modules, we design a complementary set of ablation studies (Tabs.~\ref{tab:syn_module_1}, \ref{tab:syn_module_2}, \ref{tab:syn_module_3}) focused on two core objectives: (1) whether VidRefiner preserves the intrinsic performance of \ding{172}–\ding{175} (e.g., anti-aliasing, shadow manipulation accuracy, weather effect realism); and (2) whether these combinations further boost overall spatial quality. For quantitative evaluation, we adopt PSNR (Peak Signal-to-Noise Ratio), a widely used metric for video quality assessment, to measure pixel-level similarity and quantify distortion. PSNR is computed between video samples generated by the tested module combinations and reference outputs from our full pipeline (i.e., the complete framework integrating all modules).

\begin{figure*}[t]
  \centering
    \includegraphics[width=1.0\linewidth]{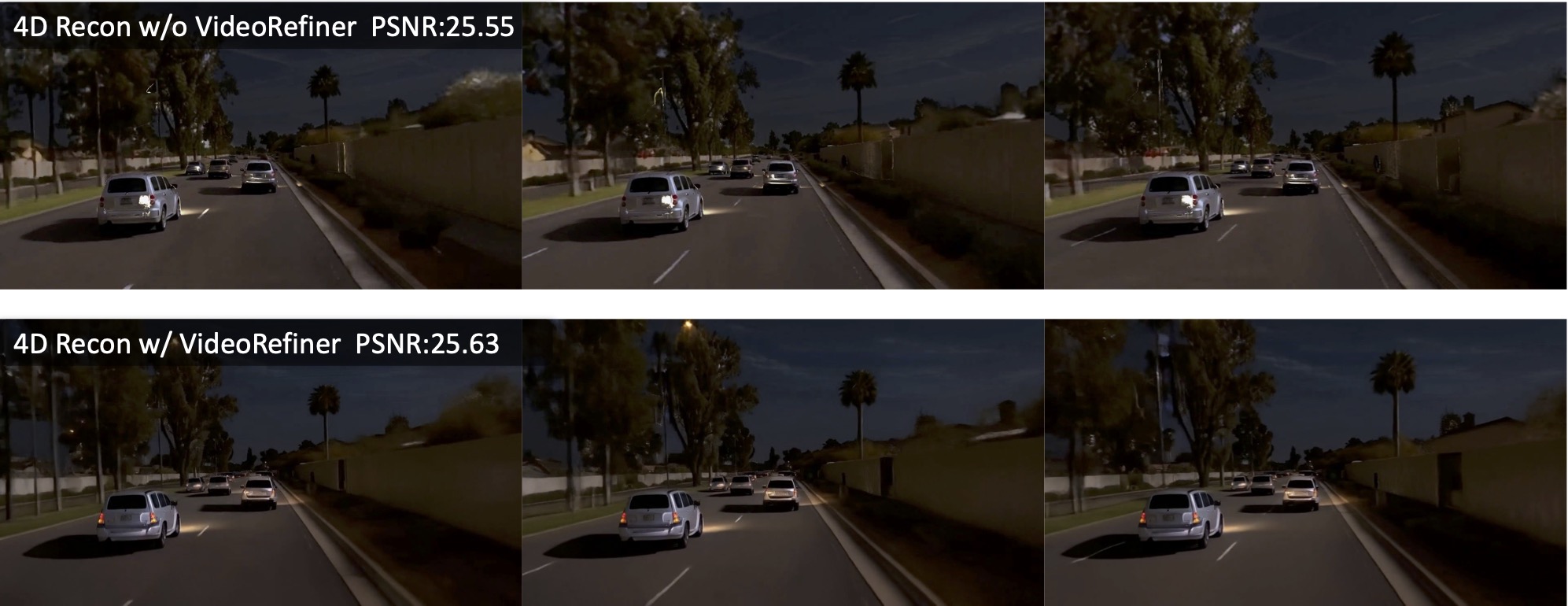}
    \caption{Experiment F: Compare 4D Reconstruction editing results with/without VidRefiner
}
    \label{fig:ab_F}
\end{figure*}

\begin{figure*}[t]
  \centering
    \includegraphics[width=1.0\linewidth]{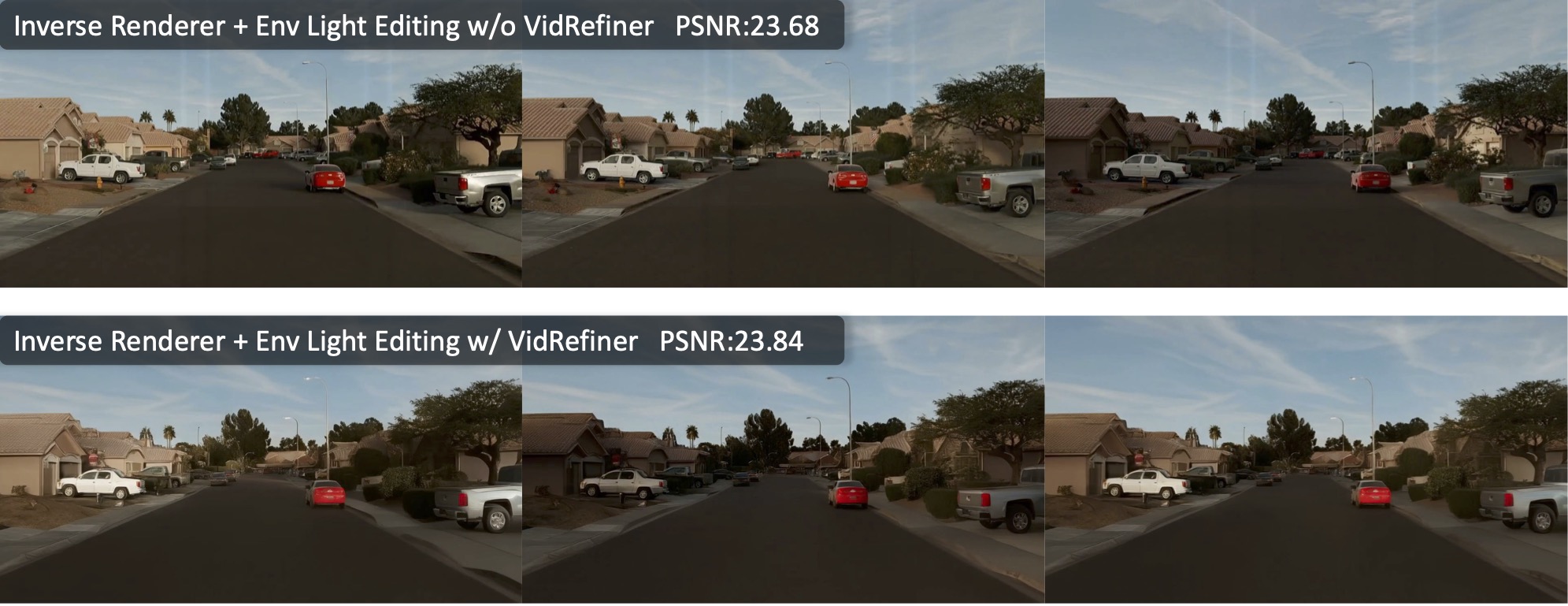}
    \caption{Experiment G: Compare Inverse Render + Light Pass editing results with/without VidRefiner.
}
    \label{fig:ab_G}
\end{figure*}

\begin{figure*}[t]
  \centering
    \includegraphics[width=1.0\linewidth]{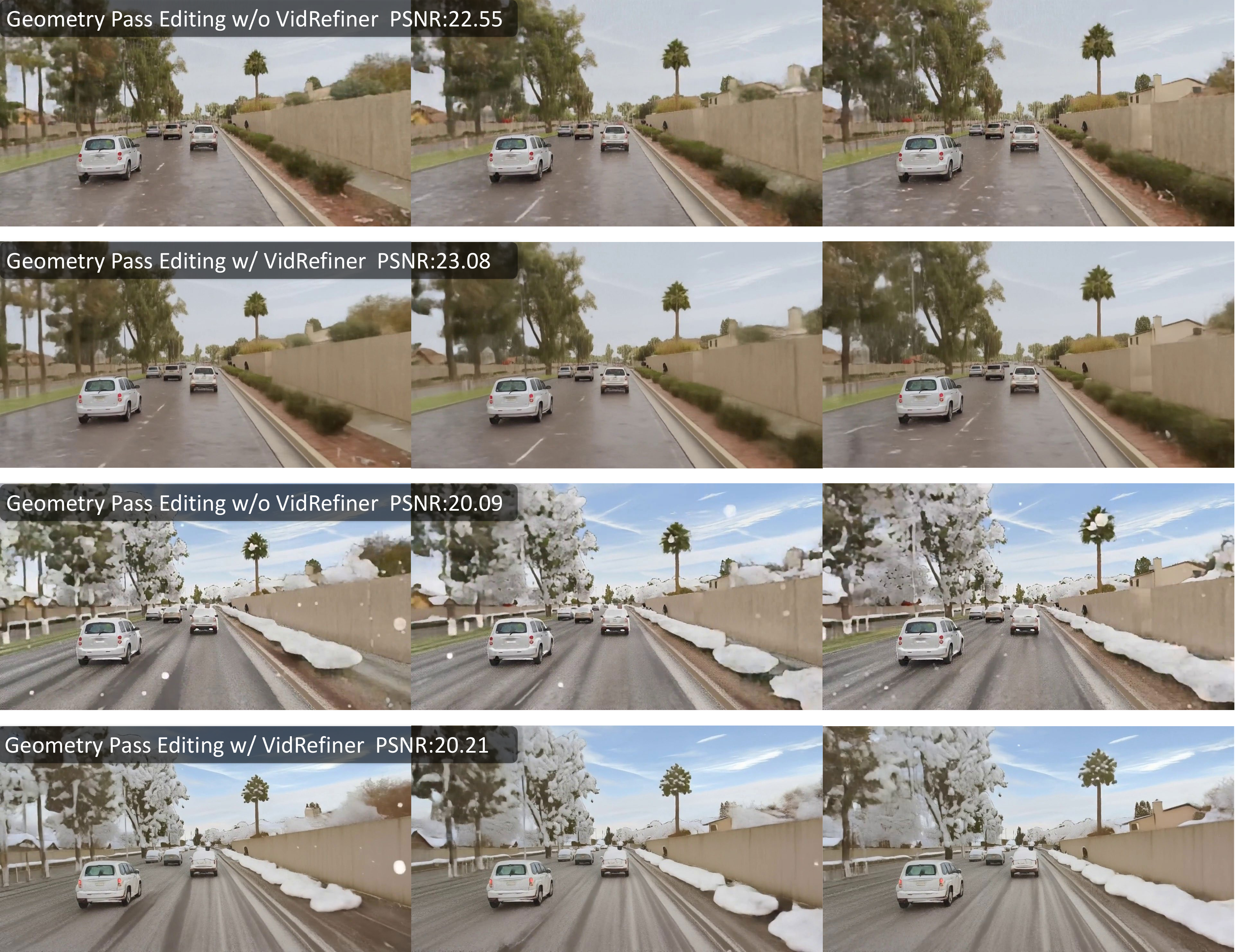}
    \caption{Experiment H: Compare rain Geometry Pass editing results with/without VidRefiner, and snow Geometry Pass editing results with/without VidRefiner.
}
    \label{fig:ab_H}
\end{figure*}

\begin{figure*}[t]
  \centering
    \includegraphics[width=1.0\linewidth]{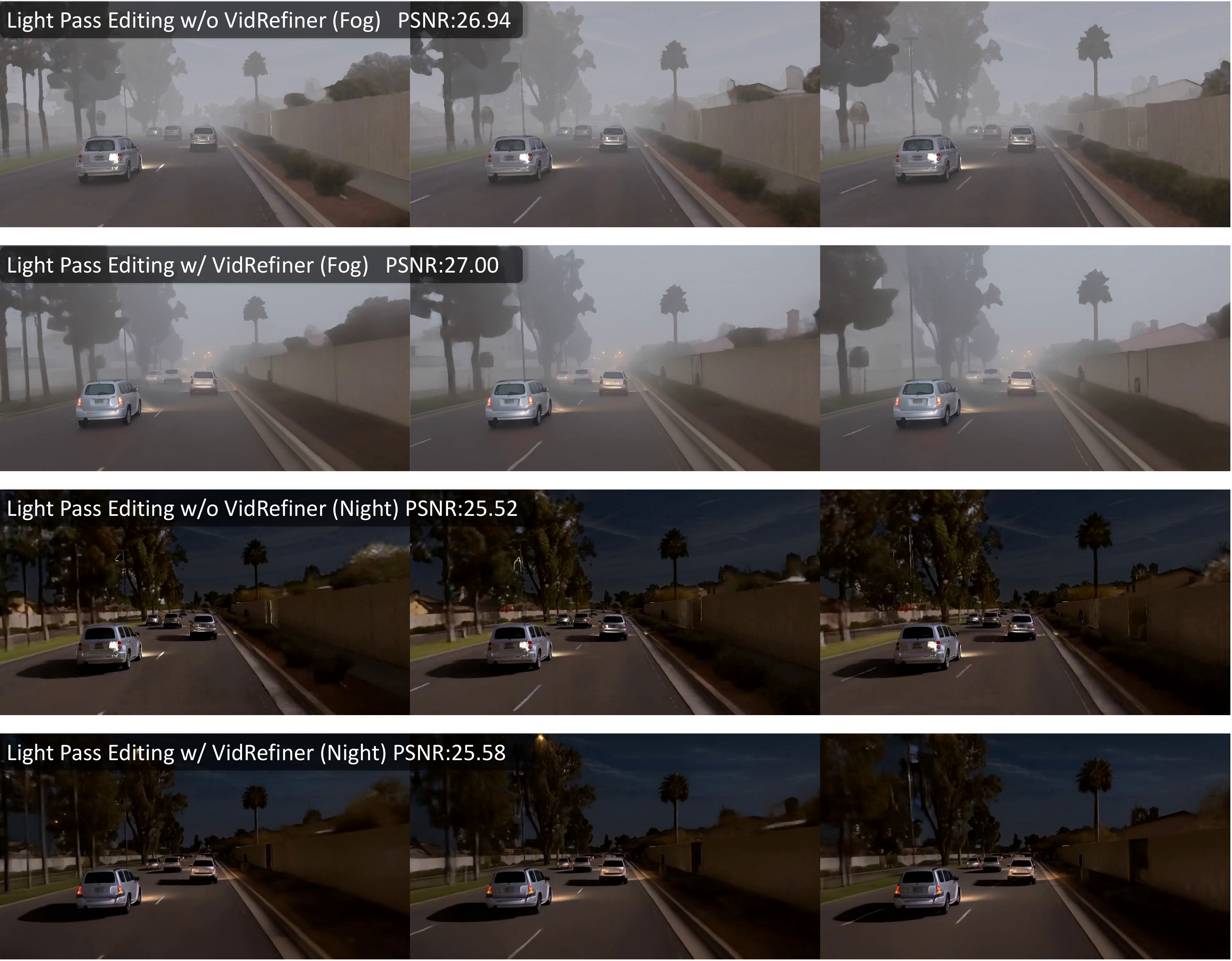}
    \caption{Experiment I: Compare fog Light Pass editing results with/without VidRefiner, and night Light Pass editing results with/without VidRefiner.
}
    \label{fig:ab_I}
\end{figure*}

\clearpage

\begin{table*}[htbp]
  \centering
  \caption{Ablation Study on Module Synergies (VidRefiner + 2-Module Subsets)}
  \label{tab:syn_module_2}
  \begin{tabularx}{\linewidth}{@{}l X X @{}}
    \toprule
    \textbf{ID Combination} & \textbf{Core Verification Point} & \textbf{Experimental Configuration} \\
    \midrule
    \ding{172}+\ding{173}+\ding{176} 
    & Anti-aliasing + shadow manipulation + fidelity 
    & Not feasible: Anti-aliasing (4D reconstruction) acts primarily on light rendering, with no direct relevance to shadow manipulation. \\
    \ding{172}+\ding{174}+\ding{176} 
    & Anti-aliasing + snow/rain editing + fidelity 
    & Not feasible: Anti-aliasing (4D reconstruction) targets light rendering artifacts, which are irrelevant to snow/rain geometry edits. \\
    \ding{172}+\ding{175}+\ding{176} 
    & Anti-aliasing + fog/night editing + fidelity 
    & Redundant (Duplicate of Experiment F): Exact same setup as Experiment F (Tab.~\ref{tab:syn_module_1}); results are reused to avoid redundant testing. \\
    \ding{173}+\ding{174}+\ding{176} 
    & Shadow manipulation + snow/rain editing + fidelity 
    & Not feasible: Snow/rain scenarios typically involve overcast conditions, where shadows are indistinct or absent, making shadow manipulation irrelevant. \\
    \ding{173}+\ding{175}+\ding{176} 
    & Shadow manipulation + fog/night editing + fidelity 
    & Not feasible: Night/fog conditions obscure direct light sources (e.g., sunlight), resulting in minimal or no distinct shadows to manipulate. \\
    \ding{174}+\ding{175}+\ding{176} 
    & Snow/rain + fog/night editing + fidelity 
    & Experiment J: Compare results of ``Geometry Pass Editing + Light Pass Editing alone'' vs. ``the combination + VidRefiner'' in Fig.~\ref{fig:ab_J}. \\
    \bottomrule
  \end{tabularx}
\end{table*}

\begin{figure*}[t]
  \centering
    \includegraphics[width=1.0\linewidth]{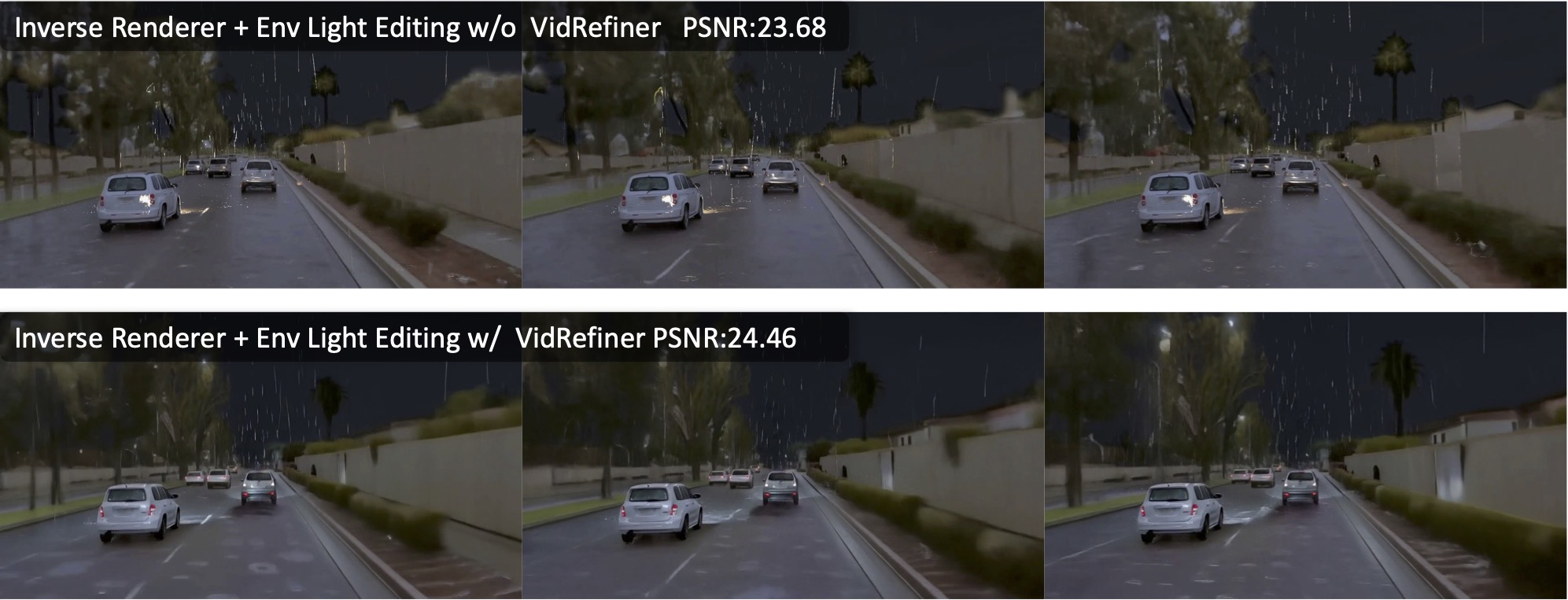}
    \caption{Experiment J: Compare Geometry Pass
Editing + Light Pass editing results with/without VidRefiner
}
    \label{fig:ab_J}
\end{figure*}

\clearpage

\begin{table*}[htbp]
  \centering
  \caption{Ablation Study on Module Synergies (VidRefiner + 3-Module Subsets)}
  \label{tab:syn_module_3}
  \begin{tabular}{@{}l p{4.5cm} p{4.5cm}@{}}
    \toprule
    \textbf{ID Combination} & \textbf{Core Verification Point} & \textbf{Experimental Configuration} \\
    \midrule
    \ding{172}+\ding{173}+\ding{174}+\ding{176} 
    & Anti-aliasing + shadow manipulation + snow/rain editing + fidelity 
    & Not feasible: Anti-aliasing (via 4D reconstruction) targets light rendering artifacts and is irrelevant to shadow manipulation or snow/rain geometry edits. \\
    \ding{172}+\ding{173}+\ding{175}+\ding{176} 
    & Anti-aliasing + shadow manipulation + fog/night editing + fidelity 
    & Not feasible: Night/fog conditions obscure direct light sources (e.g., sunlight), resulting in minimal distinct shadows, making shadow manipulation irrelevant. \\
    \ding{172}+\ding{174}+\ding{175}+\ding{176} 
    & Anti-aliasing + snow/rain + fog/night editing + fidelity 
    & Redundant (Duplicate of Experiment J): Exact same setup as Experiment J (Tab.~\ref{tab:syn_module_2}); results are reused to avoid redundant testing. \\
    \ding{173}+\ding{174}+\ding{175}+\ding{176} 
    & Shadow manipulation + snow/rain + fog/night editing + fidelity 
    & Not feasible: Night/fog conditions eliminate distinct shadows, rendering the shadow manipulation module irrelevant to combined snow/rain and fog/night weather edits. \\
    \bottomrule
  \end{tabular}
\end{table*}

\clearpage

\subsection{Intra-module Ablation Studies}

In this section, we discuss intra-module ablation studies focusing on Geometry Pass Editing and Light Pass Editing. We present the effects of their key components in geometric manipulation and lighting control, as well as their contributions to fine-grained control over the geometry and local lighting of the final video.

\begin{figure*}[t]
  \centering
    \includegraphics[width=1.0\linewidth]{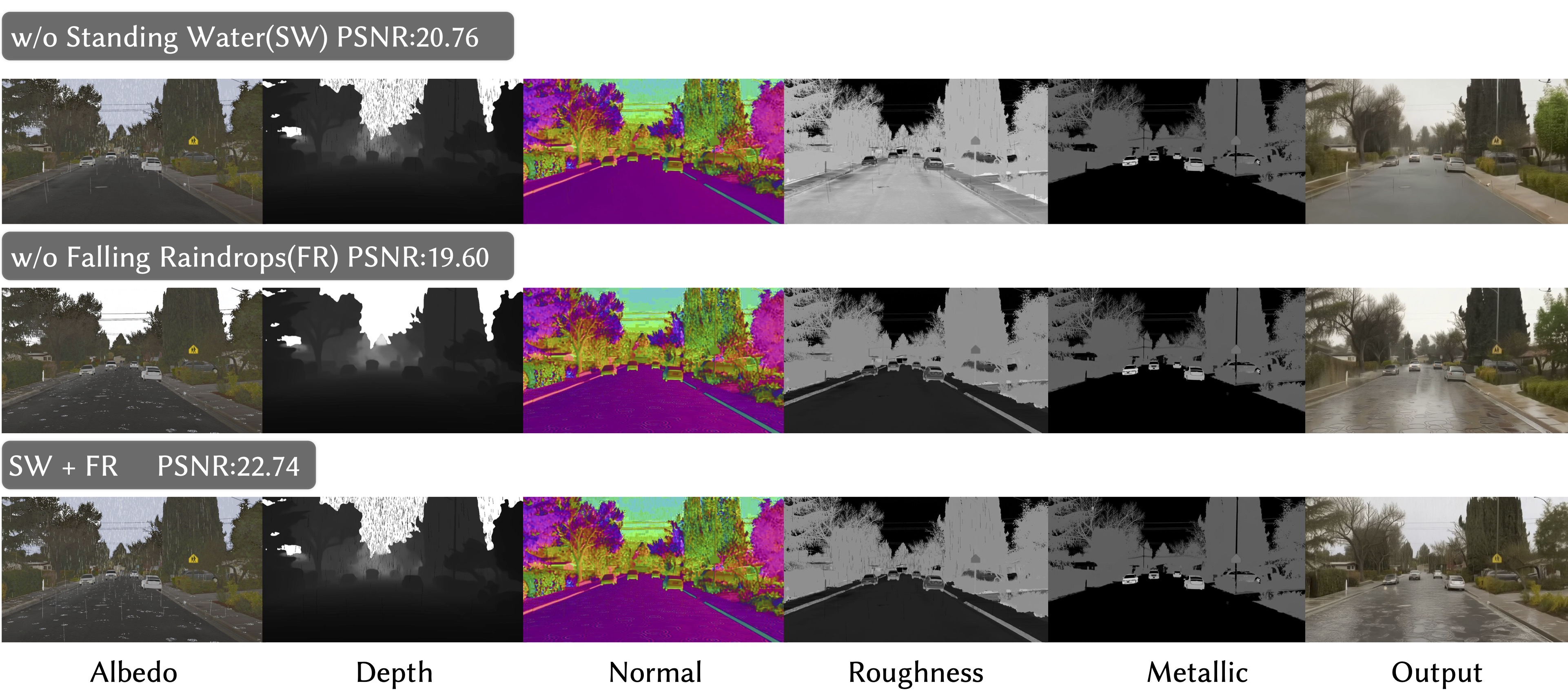}
    \caption{Comparing the individual contributions of each component in the rain-related Geometry Pass Editing.
}
    \label{fig:ab_K}
\end{figure*}

\begin{figure*}[t]
  \centering
    \includegraphics[width=1.0\linewidth]{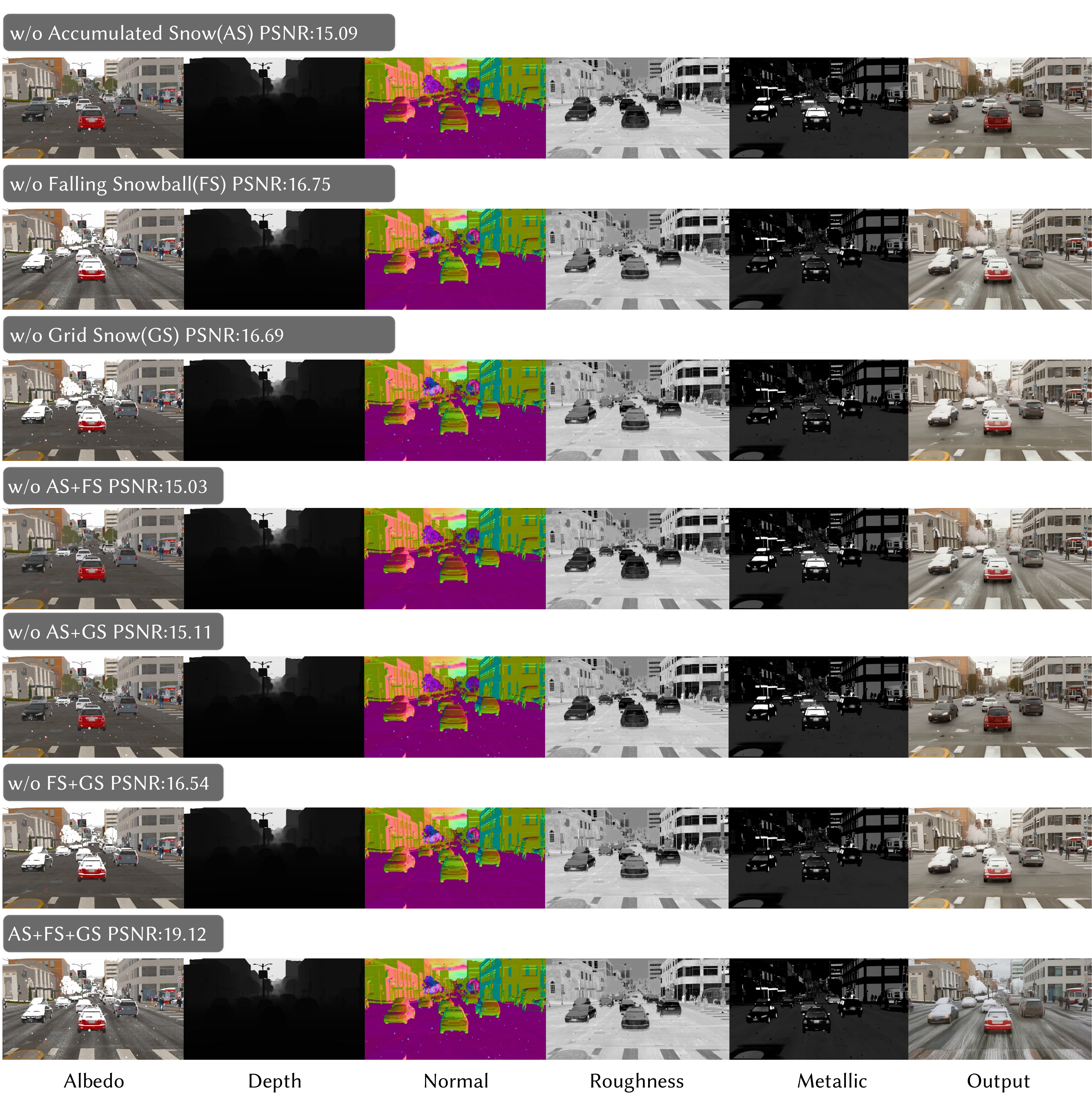}
    \caption{Comparing the individual contributions of each component in the snow-related Geometry Pass Editing.
}
    \label{fig:ab_L}
\end{figure*}

\paragraph{Geometry Pass Editing Ablation Study.} For rain-related Geometry Pass Editing, the key components are divided into two categories: standing water and falling water. We ablate these components independently to demonstrate their respective qualitative and quantitative impacts on the final weather editing results in Fig~\ref{fig:ab_K}. For snow-related Geometry Pass Editing, the core components are further partitioned into three subsets: accumulated snow, falling snowballs, and grid-based snow. The details of this ablation study are presented in Fig.~\ref{fig:ab_L}.

\begin{figure*}[t]
  \centering
    \includegraphics[width=1.0\linewidth]{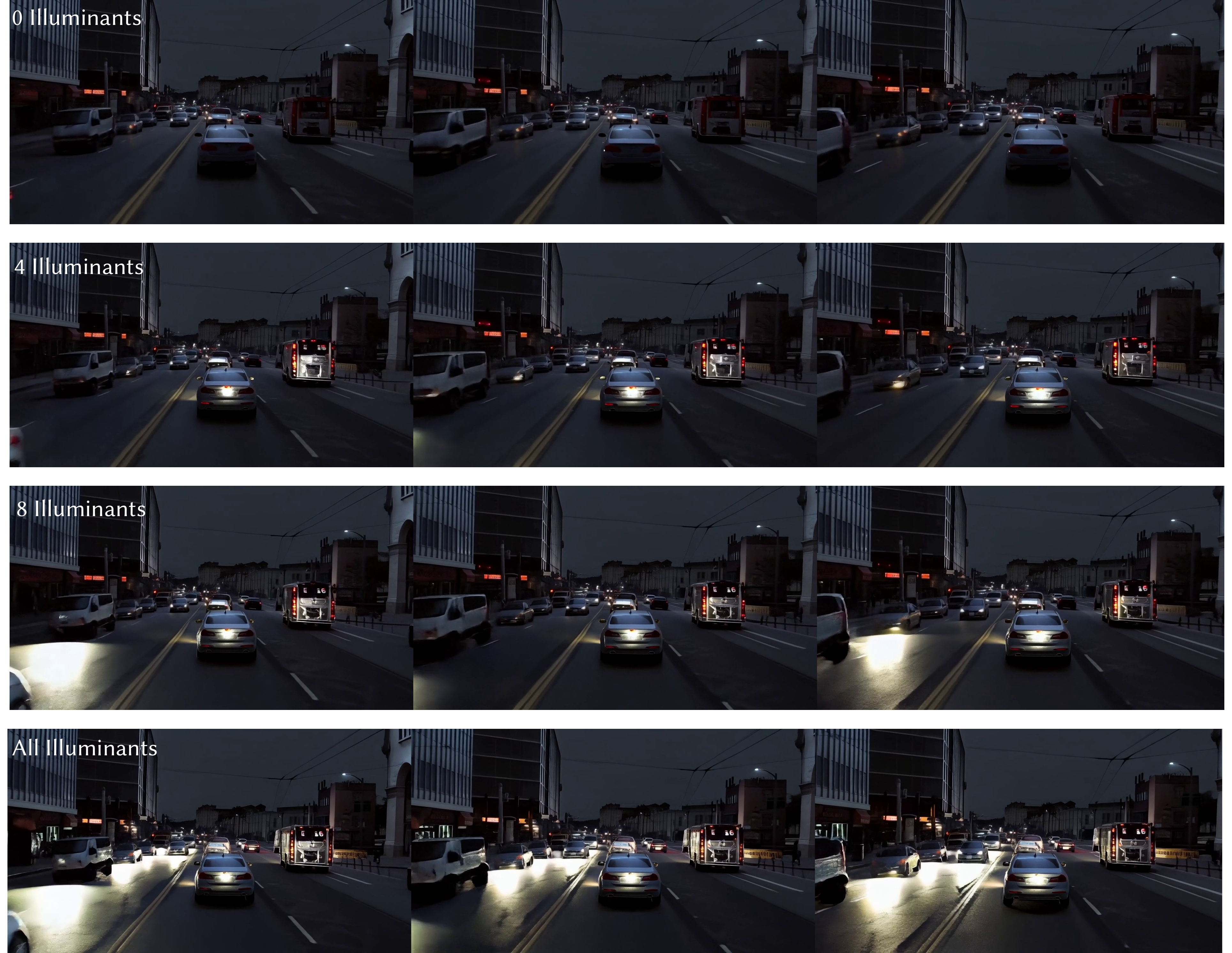}
    \caption{Comparing the individual illuminant contributions of Light Pass Editing.
}
    \label{fig:ab_M}
\end{figure*}

\paragraph{Light Pass Editing Ablation Study.} The Light Pass Editing module relies on multiple illuminants for fine-grained lighting manipulation. We conduct an ablation study by varying the number of active light sources (0, 4, 8, and all illuminants) to illustrate their individual and cumulative effects on lighting fidelity and scene photorealism. The qualitative results, depicting the progression from no illumination to full multi-source lighting, are presented in Fig.~\ref{fig:ab_M}.

\begin{figure*}[t]
  \centering
    \includegraphics[width=1.0\linewidth]{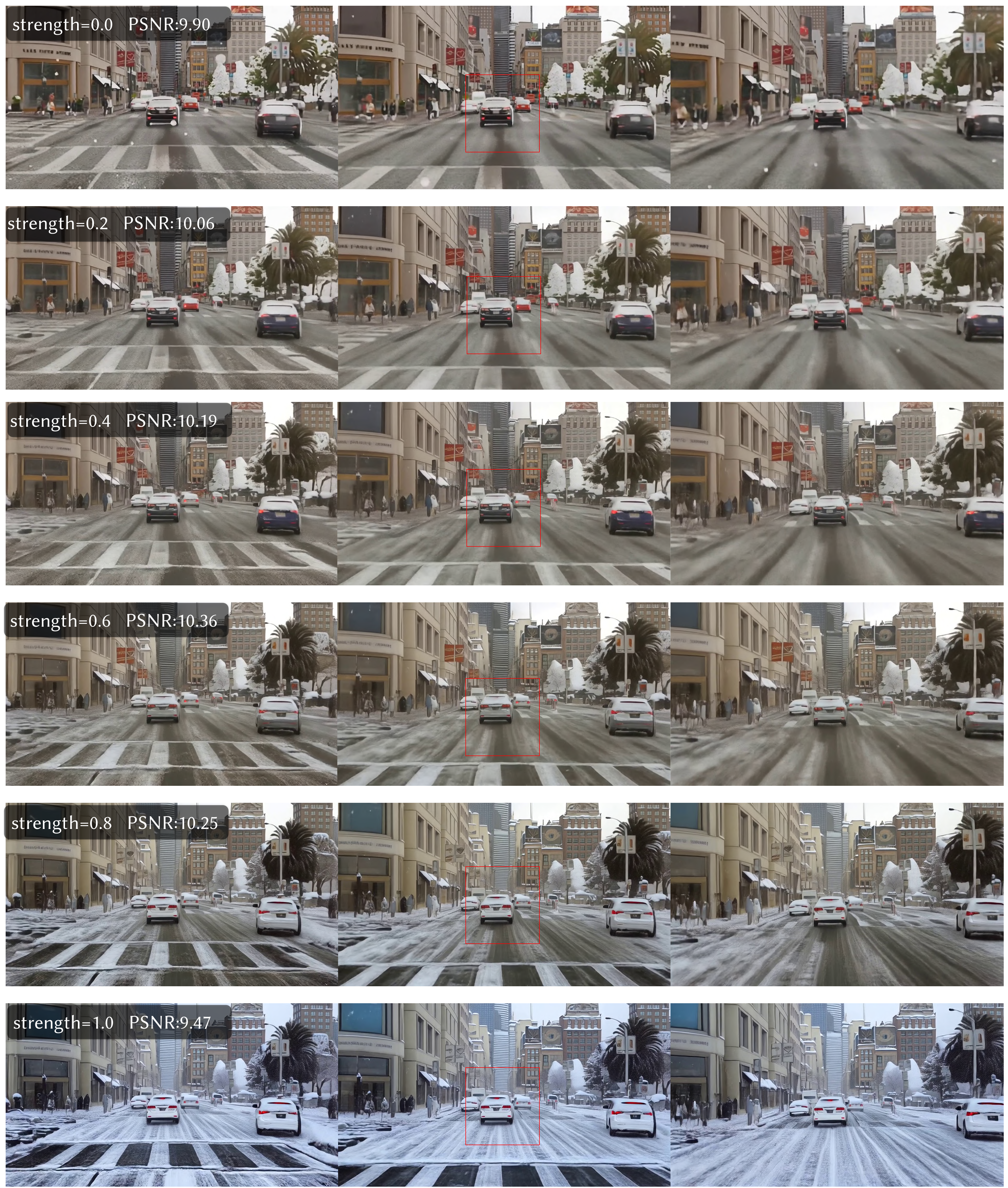}
    \caption{Ablation of VidRefiner strength across different values. 
}
    \label{fig:ab_N}
\end{figure*}

\paragraph{VidRefiner Ablation Study.} To justify our selection of VidRefiner strength as 0.4, we conduct an ablation study by varying this parameter. As illustrated in Fig.~\ref{fig:ab_N}, while a strength of 0.6 yields the highest quantitative fidelity (PSNR=10.36), it erroneously alters the car color within the red bounding box from black to white—indicating an over-strong modification tendency that deviates from physical plausibility. In contrast, a strength of 0.4 achieves a balanced trade-off: it improves visual quality (PSNR=10.19) without introducing such semantic errors, thus we adopt this value as the optimal VidRefiner strength.

\clearpage

\subsection{Case Study: Error Tolerance of G-Buffer Dual-Pass Editing}

\begin{figure*}[t]
  \centering
    \includegraphics[width=1.0\linewidth]{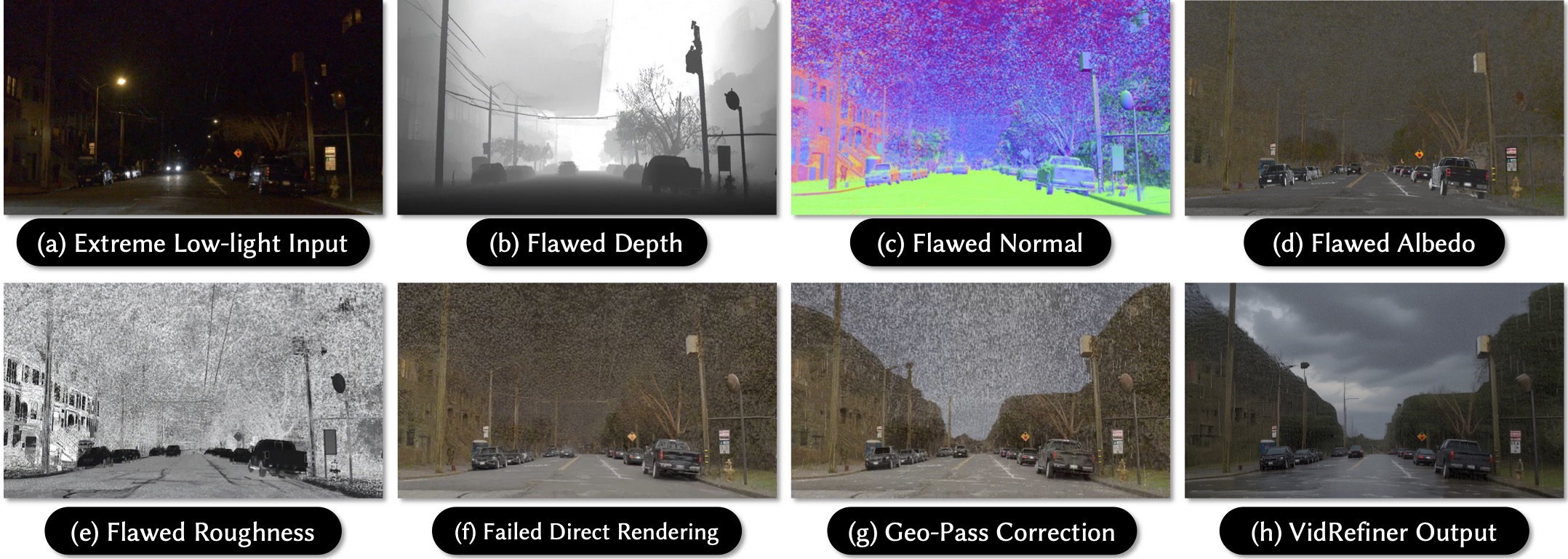}
    \caption{\textbf{Case study on error tolerance under extreme conditions.} We visualize a challenging extreme-low-light scenario to illustrate the error-compensation mechanism of our decoupled architecture. Due to high sensor ISO in the raw input (a), the feed-forward extraction yields severely flawed intrinsic components, including a catastrophic depth failure in the sky (b) and high-frequency noise in the normal/roughness maps (c, e). Naive explicit forward rendering inherently propagates these errors, leading to structural collapse (f). AutoWeather4D mitigates this through compensations: First, explicit boundary priors (e.g., sky-masking) anchor the macroscopic geometry (g); however, the intended physical weather edits on the road are shattered into high-frequency artifacts by the underlying normal noise. Second, rather than naive blurring, the generative VidRefiner treats this noisy render as a conditioning signal (h). It leverages diffusion priors to semantically absorb the intrinsic artifacts while harmonizing the fragmented weather elements into photorealistic wet-surface reflections. 
}
    \label{fig:error_tolerance}
\end{figure*}

A common critique of relying on explicit representations is the potential pipeline fragility caused by external feed-forward modules. Specifically, it is questioned how imperfect intrinsic components affect the G-Buffer Dual-Pass Editing, and whether the terminal VidRefiner compensates for or degrades the physical correctness.

To intuitively analyze this error-tolerance mechanism, we present a challenging extreme-low-light case study in Fig.~\ref{fig:error_tolerance}, where high sensor ISO induces severe intrinsic extraction failures. Rather than catastrophically propagating these errors, AutoWeather4D explicitly mitigates pipeline fragility through two compensations:

1. Explicit Macro-Structural Anchoring: Rather than blindly trusting all extracted depths, the Geometry Pass enforces explicit semantic priors (e.g., sky-masking) and metric calibration to effectively bound the physical simulation. This prevents unconstrained geometric distortion in the sky region (Fig.~\ref{fig:error_tolerance}-g).

2. Generative Error Absorption: While explicit corrections fix the macro-structure, high-frequency intrinsic noise (e.g., jagged normals) still manifests as harsh specular noise in the G-buffer output. Here, the VidRefiner acts as a crucial neural compensator. By heavily conditioning the diffusion process on this noisy but physically localized render, the model's real-world prior naturally ``absorbs'' the analytical flaws. It harmonizes the fragmented artifacts into plausible wet-surface reflections (Fig.~\ref{fig:error_tolerance}-h) without degrading the established macroscopic physical constraints. This demonstrates how our pipeline effectively breaks the chain of cascading errors.

\section{Extensive Qualitative Results}
\label{sec:more_qualitative}

\begin{figure*}[t]
  \centering
    \includegraphics[width=1.0\linewidth]{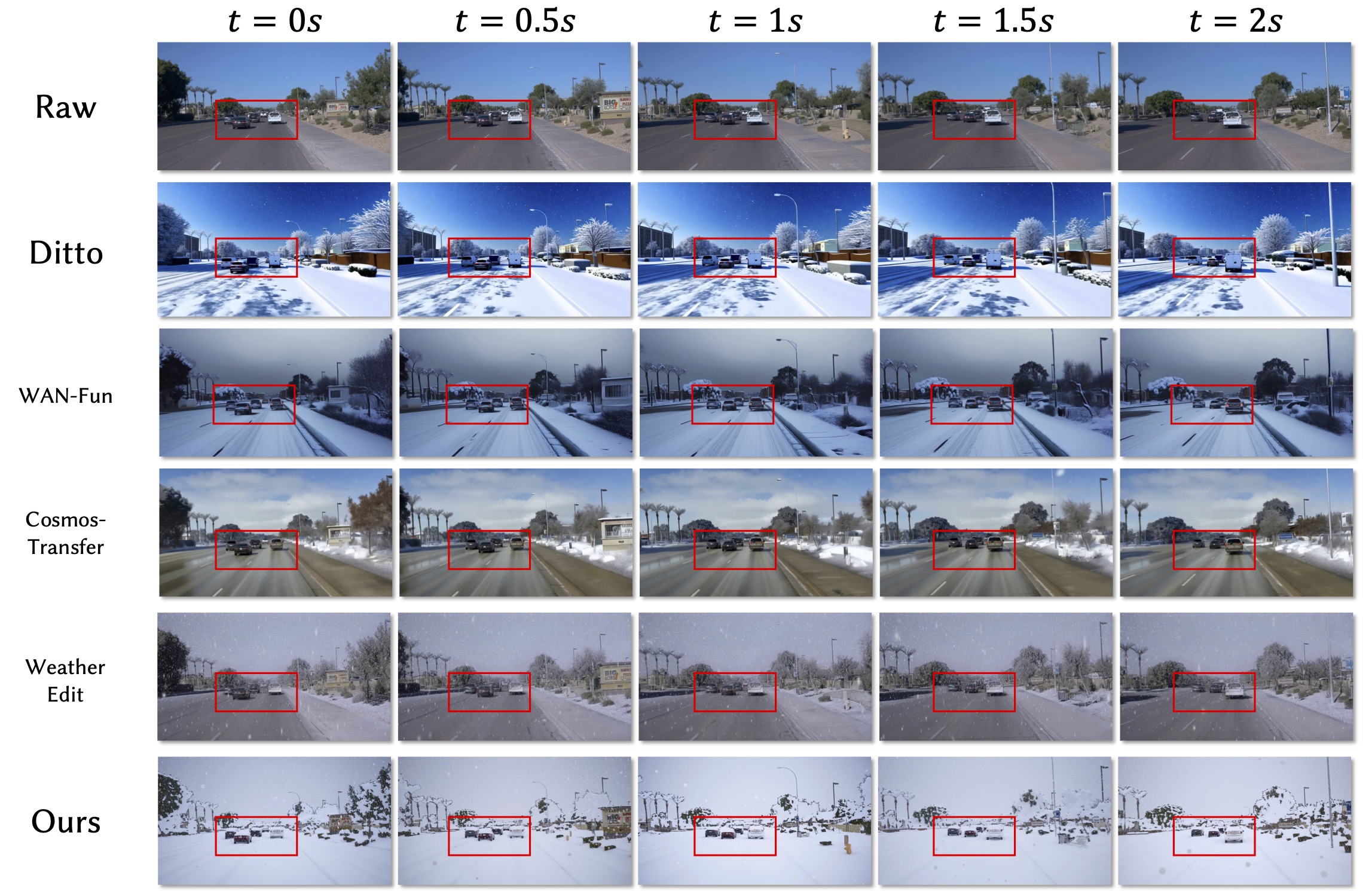}
    \caption{Qualitative comparison results. As highlighted by the red boxes, baseline models erroneously inherit hard shadows from the source video, whereas our method successfully avoids retaining these spurious shadow artifacts.}
    \label{fig:supp_qualitative}
\end{figure*}

\textbf{Qualitative Analysis: Mitigating Illumination Entanglement}
As shown in Fig.~\ref{fig:supp_qualitative}, translating scenes from sunny to snowy conditions exposes practical challenges in current baseline methods, particularly regarding the retention of original directional shadows. Physically, snowy environments are typically dominated by heavily diffused global illumination, making sharp directional shadows uncommon. This phenomenon highlights the inherent difficulty of disentangling illumination from geometry in existing approaches.

The baselines (Cosmos-Transfer, WAN-Fun, and Ditto) generally operate without explicit intrinsic decomposition. Consequently, they often struggle to differentiate between high-frequency shadow boundaries and actual structural geometry, frequently retaining these original shadows as darkened surface textures in the target domain.

Similarly, while 4DGS-based methods (e.g., WeatherEdit) effectively model atmospheric particles, their reliance on 2D image-space diffusion priors for background editing limits their ability to perform explicit illumination decoupling. As a result, the original occlusion shadows are often retained in the synthesized background rather than being diffused.

In contrast, AutoWeather4D mitigates this entanglement via explicitly decoupled G-buffers. By separating material modification (Geometry Pass) from light transport recalculation (Light Pass), our approach avoids retaining the source directional shadows. This explicit disentanglement facilitates a more physically plausible surface appearance without relying on entangled source illumination.

\begin{figure*}
  \centering
    \includegraphics[width=1.0\linewidth]{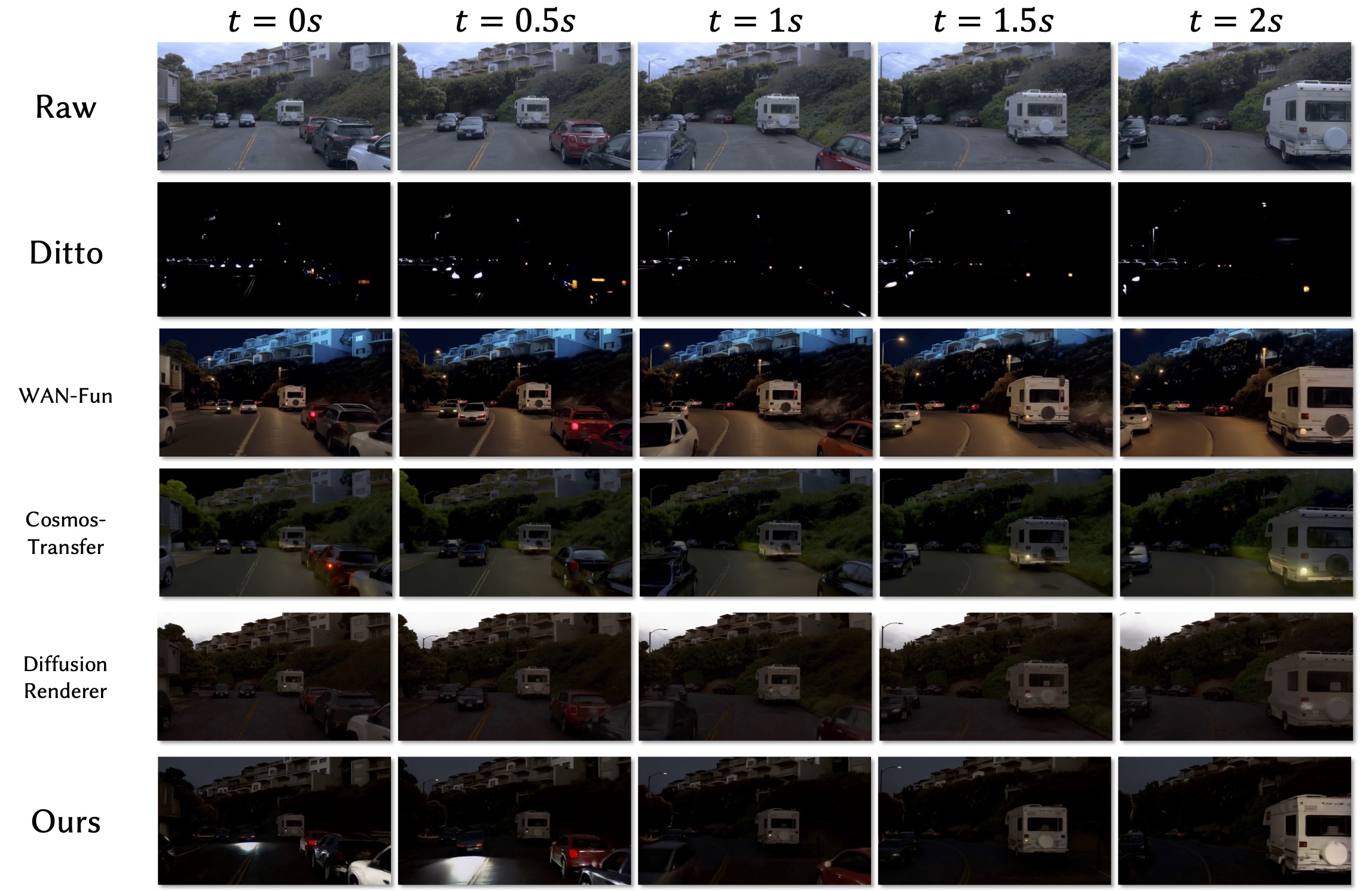}
    \caption{Qualitative comparison results. AutoWeather4D selectively synthesizes explicit headlight cones for moving vehicles, reliably maintaining the unlit appearance of the stationary parked cars. In contrast, existing baselines typically rely on global tone shifts or unstructured darkening, making it challenging to achieve such state-aware local lighting. }
    \label{fig:supp_qualitative_night}
\end{figure*}

\textbf{Qualitative Analysis: The Missing Active Illumination Problem}
As shown in Fig.~\ref{fig:supp_qualitative_night}, we observe a common artifact across existing baselines when translating scenes from day to night conditions: the difficulty in synthesizing explicit local active light sources.

Physically, nighttime driving environments are characterized by the interaction of local active illumination (e.g., ego-vehicle headlights) with the 3D road geometry, while unlit background areas remain in shadow. This highlights a limitation in current methods:

WAN-Fun and Cosmos-Transfer can synthesize a general nighttime atmosphere but often lack physical constraints for light transport. They tend to introduce global color shifts (e.g., blue tints from WAN-Fun) over the entire scene rather than projecting explicit headlight cones onto the asphalt.

Global darkening approaches (e.g., Ditto, Diffusion Renderer) primarily rely on tone mapping to uniformly darken the scene. Without explicitly modeling active light sources, they struggle to provide the necessary local illumination expected in a realistic driving scenario.

In contrast, AutoWeather4D addresses this via a decoupled Light Pass. By explicitly injecting volumetric headlight cones into the 3D space, our approach synthesizes plausible light falloff and specular reflections on the road surface and leading vehicles. Meanwhile, unlit background structures are maintained as dark silhouettes, facilitating physically grounded nighttime synthesis without unintended ambient artifacts.

\begin{figure*}
  \centering
    \includegraphics[width=1.0\linewidth]{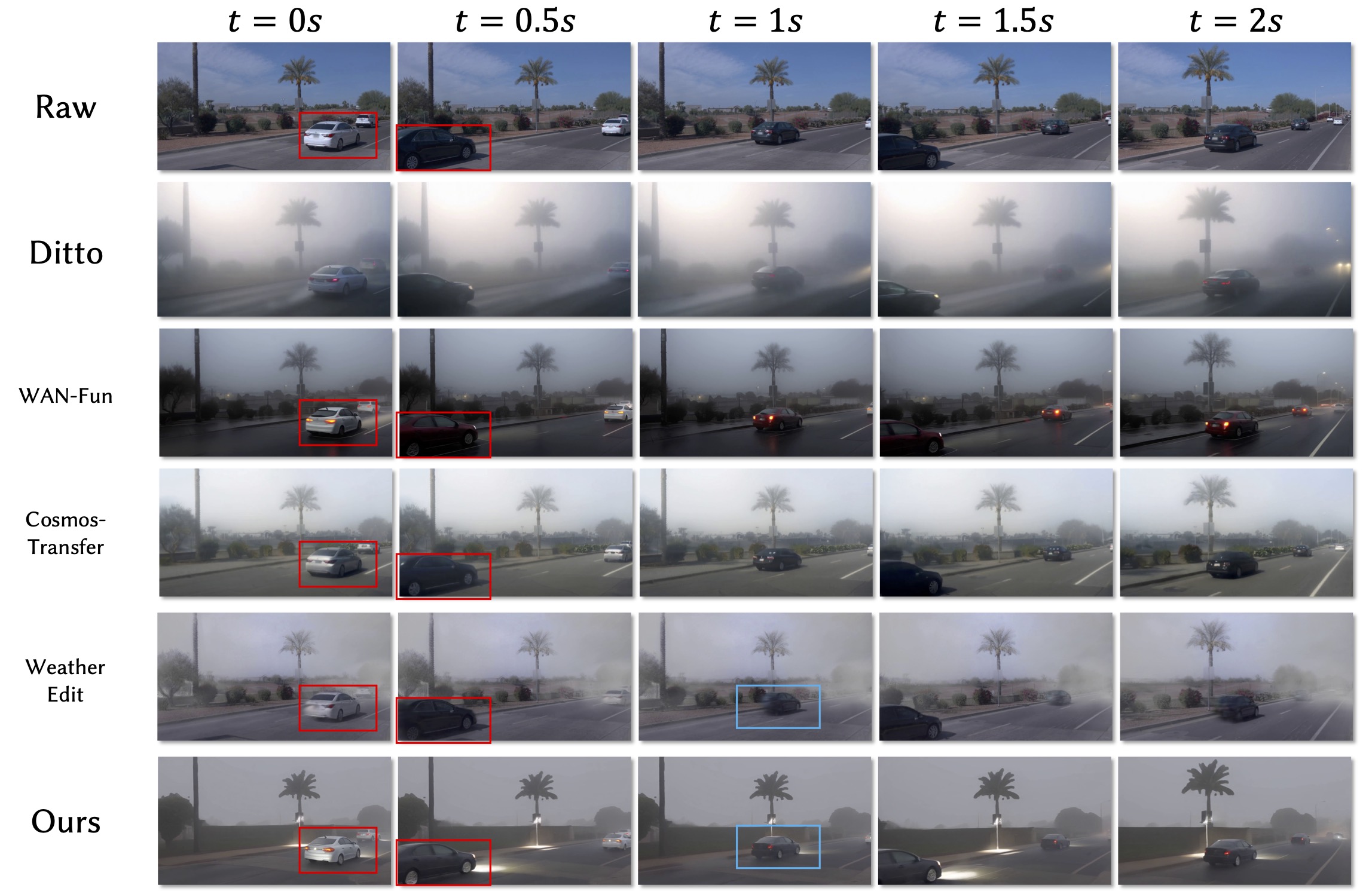}
    \caption{Qualitative comparison results. As highlighted by the red boxes, existing baselines erroneously inherit hard directional shadows from the source video, whereas our method successfully avoids this artifact and injects plausible local headlight illumination. Furthermore, the blue boxes demonstrate that Gaussian-Splatting-based methods (e.g., WeatherEdit) struggle to reconstruct dynamic moving objects from monocular videos, resulting in severe motion ghosting. In contrast, our approach preserves the structural integrity of dynamic vehicles.}
    \label{fig:supp_qualitative_fog}
\end{figure*}

\textbf{Qualitative Analysis: Shadow Inheritance and Dynamic Reconstruction in Fog}
As illustrated in Fig.~\ref{fig:supp_qualitative_fog}, synthesizing physically plausible fog from sunny driving videos exposes fundamental limitations in existing baselines across illumination handling and dynamic reconstruction.

Physically, dense fog heavily scatters sunlight into uniform ambient illumination and mandates the use of vehicle headlights due to low visibility. However, existing baselines fail on both fronts. WAN-Fun and  Cosmos-Transfer erroneously inherit the sharp occlusion shadows from the sunny source video (highlighted by red boxes). Conversely, global translation approaches like Ditto merely apply a 2D opacity filter, completely failing to synthesize the necessary active local illumination (i.e., headlights). In contrast, AutoWeather4D recalculates the light transport via the decoupled Light Pass, successfully erasing the spurious shadows while explicitly injecting physically plausible volumetric headlights.

Beyond lighting, rendering adverse weather requires robust geometric handling. As shown in the blue boxes, 4D-Gaussian-based methods (e.g., WeatherEdit) struggle to reconstruct fast-moving dynamic objects from monocular inputs. The optimization of dynamic Gaussian splats often collapses, resulting in severe motion ghosting and blurred vehicle geometries. Conversely, our feed-forward G-buffer extraction relies on robust depth tracking and avoids monolithic 4D optimization, thereby strictly preserving the structural integrity and sharp boundaries of dynamic entities.

\begin{figure*}
  \centering
    \includegraphics[width=1.0\linewidth]{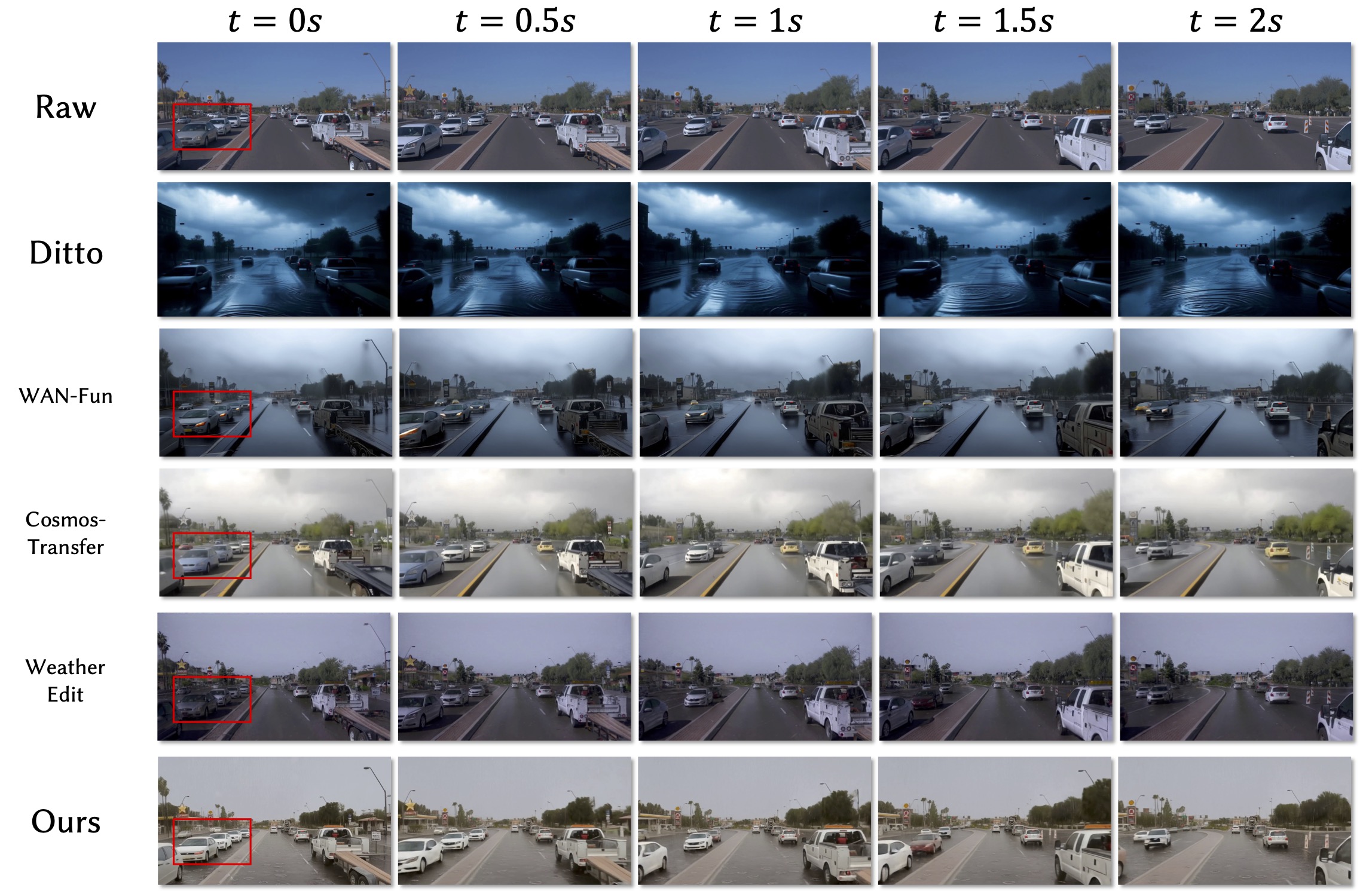}
    \caption{Qualitative comparison results. As highlighted by the red boxes, several baselines erroneously inherit hard directional shadows from the source video. }
    \label{fig:supp_qualitative_rain}
\end{figure*}

\textbf{Qualitative Analysis: Shadow Inheritance and Active Precipitation in Rain}
As shown in Fig.~\ref{fig:supp_qualitative_rain}, translating sunny scenes to rainy conditions exposes critical challenges in current baseline methods, particularly regarding illumination decoupling, and active surface interaction.

Illumination Entanglement (Red Boxes): An overcast rainy sky dictates heavily diffused ambient lighting. However, current video editing baselines (e.g., WAN-Fun, Cosmos-Transfer, Weather Edit) often struggle to decouple the original lighting from the scene geometry. As highlighted by the red boxes, they erroneously inherit the sharp directional car shadows from the sunny source video. In contrast, AutoWeather4D explicitly recalculates the light transport via the Light Pass, successfully circumventing these spurious shadows to yield physically plausible diffuse ground illumination.

Active Precipitation vs. Global Tone Shifts: Synthesizing a realistic rainy environment requires modeling active precipitation and dynamic surface interactions (e.g., puddles and ripples). Baselines like WAN-Fun, Cosmos-Transfer, and WeatherEdit primarily utilize global color temperature shifts and static road darkening. Consequently, they tend to synthesize a "post-rain" (wet road) appearance rather than an ongoing rainstorm. While Ditto attempts to generate water ripples, it struggles with maintaining strict spatial constraints, leading to noticeable distortion of the original scene layout. Conversely, our approach utilizes explicit world-space procedural modeling (Geometry Pass) to synthesize dynamic ripples on geometry-anchored puddles, ensuring the rendering of active precipitation while maintaining the structural integrity of the driving scene.

\begin{figure*}
  \centering
    \includegraphics[width=1.0\linewidth]{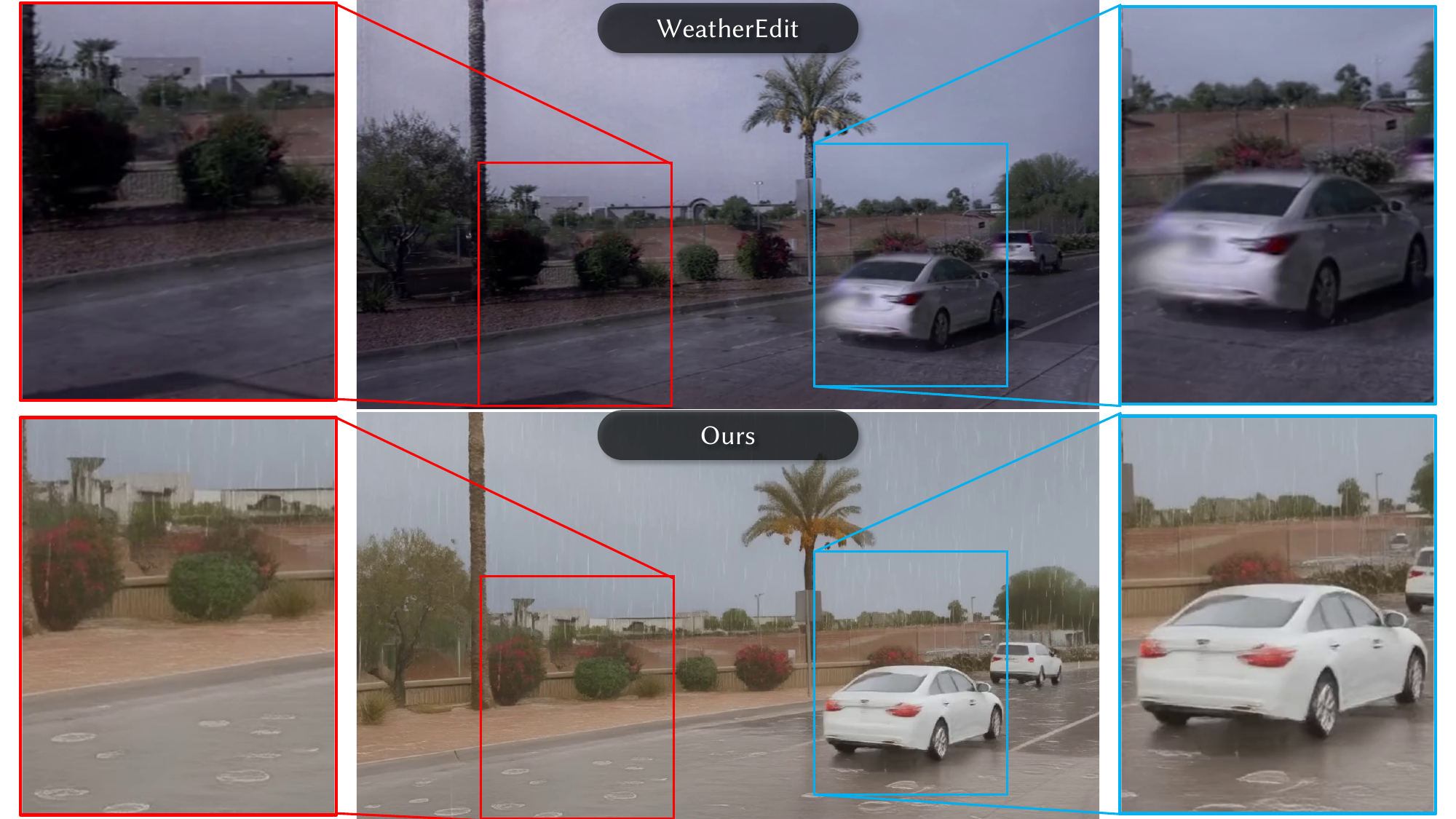}
    \caption{Qualitative comparison demonstrating the advantages of decoupled G-buffers over 4D Gaussian Splatting (WeatherEdit). (Red boxes) AutoWeather4D synthesizes explicit surface interactions (e.g., geometry-anchored puddles with dynamic ripples). In contrast, WeatherEdit struggles to apply fine-grained, geometry-aware physical modifications to the road surface. (Blue boxes) Our feed-forward G-buffer formulation maintains the structural integrity of fast-moving vehicles, effectively mitigating the motion ghosting observed in the optimization-based 4DGS baseline. }
    \label{fig:supp_detail_comparison}
\end{figure*}

\textbf{Qualitative Analysis: Dynamic Structure Preservation and Surface Interaction}
As shown in Fig.~\ref{fig:supp_detail_comparison}, applying complex weather effects to highly dynamic driving scenes exposes critical structural challenges in current 3D-aware baselines (e.g., WeatherEdit), particularly regarding motion reconstruction and fine-grained surface modifications.

\textbf{Mitigating Dynamic Reconstruction Artifacts (Blue Boxes):} Reconstructing moving objects from video remains a practical hurdle for optimization-based scene representations like 4DGS. As highlighted by the blue boxes, the 4DGS-based baseline, WeatherEdit, exhibits motion ghosting and geometric instability when processing the moving white vehicle. In contrast, AutoWeather4D extracts explicit G-buffers via a feed-forward neural network, accommodating dynamic objects. Bypassing per-scene optimization mitigates these geometric vulnerabilities, thereby helping to preserve the structural boundaries of dynamic entities.

\textbf{Disentanglement for Surface Interactions (Red Boxes):} Synthesizing interactive weather elements, such as accumulated puddles with dynamic ripples, generally relies on explicit geometric grounding. While WeatherEdit models atmospheric particles in 3D, its reliance on 2D image-space diffusion priors for background editing limits its ability to perform explicit intrinsic decomposition. Consequently, it struggles to support geometry-aware, localized water dynamics on the road surface. Conversely, our approach explicitly decouples geometry and illumination, providing a controllable geometric foundation. This disentanglement enables the Geometry Pass to directly modulate intrinsic material properties—synthesizing geometry-anchored puddles and perturbing surface normals for ripples—facilitating physically plausible effects that remain challenging for existing 4DGS-based works.

\begin{figure*}
  \centering
    \includegraphics[width=1.0\linewidth]{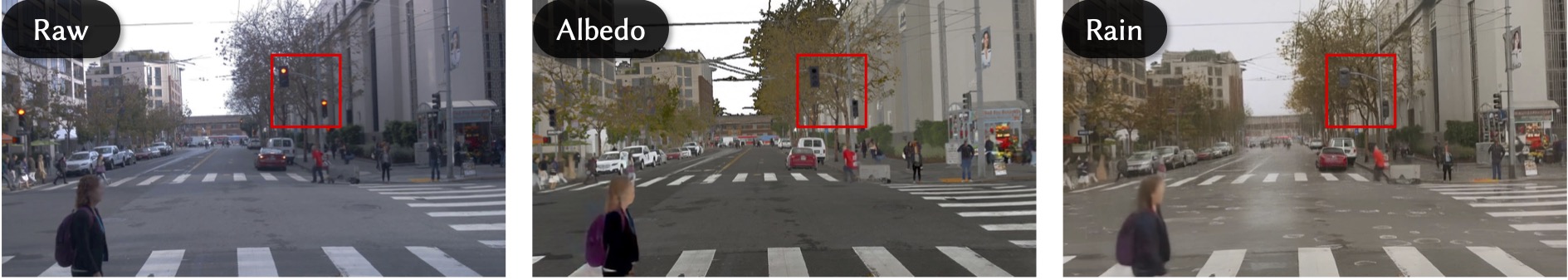}
    \caption{\textbf{Limitation in handling light-emitting objects.} To maintain 3D reconstruction stability, our G-buffer design deliberately omits a specific channel for self-illuminating objects. Consequently, active light sources like traffic lights (red boxes) are modeled as high-albedo reflective surfaces. As a result, they experience unintended darkening when the global sunlight is reduced during sunny-to-rainy translation.}
    \label{fig:failure_case}
\end{figure*}

\textbf{Failure Case in Preserving Light-Emitting Objects:}
Our decoupled G-buffer approach works well for standard surfaces, but it remains challenging to preserve the brightness of self-illuminating objects, such as traffic lights. As shown in Fig.~\ref{fig:failure_case}, when converting a bright sunny scene to a gloomy rainy environment, the brightly colored traffic signals lose their original glow and become unexpectedly dark.

This phenomenon is the result of a deliberate design choice in our inverse rendering process. To make our feed-forward extraction stable and reliable, the current G-buffer focuses on standard physical properties but intentionally leaves out a specific emissive channel. To compensate for this missing channel, the network is forced to bake the traffic light's brightness into its albedo. The system essentially treats the traffic light as a passive, high-albedo reflective surface rather than an active light emitter. Because albedo relies entirely on external illumination to be visible, these surfaces naturally go dark when the strong direct sunlight is replaced by diffused overcast lighting during the rain synthesis.

In the future, AutoWeather4D can be improved by adding a dedicated emissive channel to the G-buffer. We could utilize domain-specific traffic signal detectors or VLMs to explicitly locate these objects and maintain their brightness, which will further benefit the perception safety of autonomous driving.

%
%

\end{document}